\documentclass{nature}
\usepackage[ruled,vlined]{algorithm2e}
\usepackage{amssymb}
\usepackage{xcolor}
\usepackage[utf8]{inputenc}
\usepackage{amsmath}
\usepackage{placeins}
\usepackage{float} 

\date{\today}

\newcommand{\citeasnoun}[1]{Ref.~\citenum{#1}}
\newcommand{\secref}[1]{Sec.~\ref{#1}}

\renewcommand{\eqref}[1]{Eq.~(\ref{eq:#1})}

\newcommand{\figref}[1]{Fig.~\ref{#1}}

\newcommand{\edit}[1]{{\color{black} #1}}
\usepackage{graphicx}
\usepackage{url}

\usepackage{filecontents}

\title{Physics-enhanced deep surrogates for PDEs} 
\author{Rapha{\"e}l~Pestourie$^{1,\ast}$, Youssef~Mroueh$^{2,3}$, Chris~Rackauckas$^{4}$, Payel~Das$^{2,\ast}$ \& Steven~G.~Johnson$^4$}

\date{}

\date{\today}

\begin{document}

\maketitle

\noindent \normalsize{$^{1}$ School of Computational Science and Engineering, Georgia Institute of Technology, Atlanta, GA 30332, USA} \\
\normalsize{$^{2}$ IBM Research AI, IBM Thomas J Watson Research Center,  Yorktown Heights, NY 10598, USA}\\
\normalsize{$^{3}$ MIT-IBM Watson AI Lab, Cambridge, MA 02139, USA}\\
\normalsize{$^{4}$ MIT, Cambridge, MA 02139, USA}\\
\normalsize{$^\ast$Correspondence to: rpestourie3@gatech.edu; daspa@us.ibm.com.}

\begin{abstract}
Many physics and engineering applications demand Partial Differential Equations (PDE) property evaluations that are traditionally computed with resource-intensive high-fidelity numerical solvers. Data-driven surrogate models provide an efficient alternative but come with a significant cost of training. Emerging applications would benefit from surrogates with an improved accuracy–cost tradeoff, while studied at scale. Here we present a ``physics-enhanced deep-surrogate'' (``PEDS'') approach towards developing fast surrogate models for complex physical systems, which is described by PDEs. Specifically, a combination of a low-fidelity, explainable physics simulator and a neural network generator is proposed, which is trained end-to-end to globally match the output of an expensive high-fidelity numerical solver. Experiments on three exemplar testcases, diffusion, reaction--diffusion, and electromagnetic scattering models, show that a PEDS surrogate can be up to 3$\times$ more accurate than an ensemble of feedforward neural networks with limited data ($\approx 10^3$ training points), and reduces the training data need by at least a factor of 100 to achieve a target error of 5\%. Experiments reveal that PEDS provides a general, data-driven strategy to bridge the gap between a vast array of simplified physical models with corresponding brute-force numerical solvers modeling complex systems, offering accuracy, speed, data efficiency, as well as physical insights into the process. 
\end{abstract}

\section{Introduction}

In mechanics, optics, thermal transport, fluid dynamics, physical chemistry, climate models, crumpling theory, and many other fields, data-driven surrogate models---such as polynomial fits, radial basis functions, or neural networks---are widely used as an efficient solution to replace repetitive calls to slow numerical solvers~\cite{baker2019workshop, benner2015survey, willard2020integrating, hoffmann2019machine, pant2021deep, pestourie2018inverse}. 
However the reuse benefit of surrogate models comes at a significant training cost, in which a costly high-fidelity numerical solver must be evaluated many times to provide an adequate training set, and this cost rapidly increases with the number of model parameters (the ``curse of dimensionality'')~\cite{boyd2007chebyshev}. 
In this paper, we explore one promising route to increasing training-data efficiency: incorporating \emph{some} knowledge of the underlying physics into the surrogate by training a generative neural network (NN) ``end-to-end'' with an \emph{approximate} physics model.  We call this hybrid system a ``physics-enhanced deep surrogate'' (PEDS). 
We demonstrate multiple-order-of-magnitude improvements in sample and time complexity on three exemplar test problems involving the diffusion equation's flux, the reaction-diffusion equation's flux, and Maxwell's-equations' complex transmission coefficient for optical metamaterials---composite materials whose properties are designed via microstructured geometries~\cite{pestourie2020active}, while conforming to general robustness (SI, General robustness). 
In inverse design (large-scale optimization) of nanostructured thermal materials, chemical reactors, or optical metamaterials, the same surrogate model capturing important geometric aspects of the system may be re-used thousands or millions of time~\cite{lu2022multifidelity,pestourie2018inverse, pestourie2020assume}, making surrogate models especially attractive to accelerate computational design~\cite{bayati2021inverse, li2021inverse}.

To obtain an accurate surrogate of a PDE, we apply a deep NN to \emph{generate a low-fidelity geometry, optimally mixed with the downsampled geometry}, which is then used as an input into an approximate low-fidelity solver and trained end-to-end to minimize the overall error, as depicted in Fig.~\ref{fig:PEDS_diagram} (Sec.~\ref{sec:results}).  The low-fidelity solver may simply be the same numerical method as the high-fidelity PDE solver except at a lower spatial resolution, or it may have additional simplifications in the physics (as in the reaction--diffusion example below, where the low-fidelity model discards the nonlinear term of the PDE). When only real-world data is available, the low-fidelity model may come from a guess.  By design, this low-fidelity solver
yields unacceptably large errors in the target output (perhaps $> 100\%$), but it is orders of magnitude faster than the high-fidelity model while qualitatively preserving at least some of the underlying physics. We provide an inclusion criterion for the choice of low-fidelity model that guarantees a universal approximation and a methodology that prevents PEDS from degrading the low-fidelity solver. The NN is trained to nonlinearly correct for these errors in the low-fidelity model, but the low-fidelity model ``builds in'' some knowledge of the physics and geometry that improves the data efficiency of the training. For example, the low-fidelity diffusion model enforces conservation of mass, while the low-fidelity Maxwell model automatically respects conservation of energy and reciprocity~\cite{potton2004reciprocity},  and we can also enforce geometric symmetries; all of these augment the ``trustworthiness''~\cite{li2021trustworthy} of the model. Compared to a NN-only baseline model (SI, Implementation details of PEDS and baseline), 
we find that, with a very small dataset of $\approx 1000$ points and for several very different PDEs, PEDS consistently increases the accuracy by up to 3$\times$ compared to the NN baseline, and reduces the need for training data by an order of magnitude. For the number of parameters of the surrogate models we tested, it amounts to a Cartesian product of less than two points in each input direction. To obtain a $\approx5$\% error, comparable to fabrication uncertainty, PEDS reduces the data need by a factor of at least 100 compared to competing approaches. In the more challenging case of our surrogate of the complex optical transmission, PEDS seems to improve the asymptotic \emph{rate} of learning ($\approx 5\times$ larger power law), so that the benefits increase as accuracy tolerance is lowered (Fig.~\ref{fig:resultfigure} and \secref{sec:results}). We show through an ablation study of the surrogate for Maxwell's equations that adding information from the downsampled structure increases the accuracy by 15\% in a low-data regime. 
We extensively compared PEDS against traditional surrogate models and find that they outperforms all models for more complex problems (SI, Comparison to mainstream surrogate models).
Furthermore, when the low-fidelity solver layer is very inaccurate, we find that PEDS gains significant additional benefits by combining it with active-learning techniques from our earlier work~\cite{pestourie2020active}, and in fact the benefits of active learning (AL) seem to be even greater for PEDS than for competing approaches.  Although the resulting PEDS surrogate is more expensive to evaluate than a NN by itself due to the low-fidelity solver, it is still much faster than the high-fidelity solver with two to four orders of magnitude speedup. Furthermore, since the NN generates a downsampled version of the geometry, this output can be further examined to gain insight into the fundamental nonlinear physical processes captured by the low-fidelity solver.

\section{Results}
\subsection{PEDS general approach}
\label{sec:results}
The PEDS surrogate model $\tilde{f}(p)$ aims to predict $f^{hf}(\mathrm{hf}(p))$---an output property of interest as it would be computed from a computationally intensive high-fidelity (hf) solver $f^{hf}$. The hf solver computes the PDE solution for a high-fidelity geometry $\mathrm{hf}(p)$, with $p$ being some parameterization of the geometry (or other system parameters). PEDS is depicted schematically in~\figref{fig:PEDS_diagram}, and is implemented in the following stages:  

\begin{figure}[ht!]
    \centering
    \includegraphics[width=\textwidth]{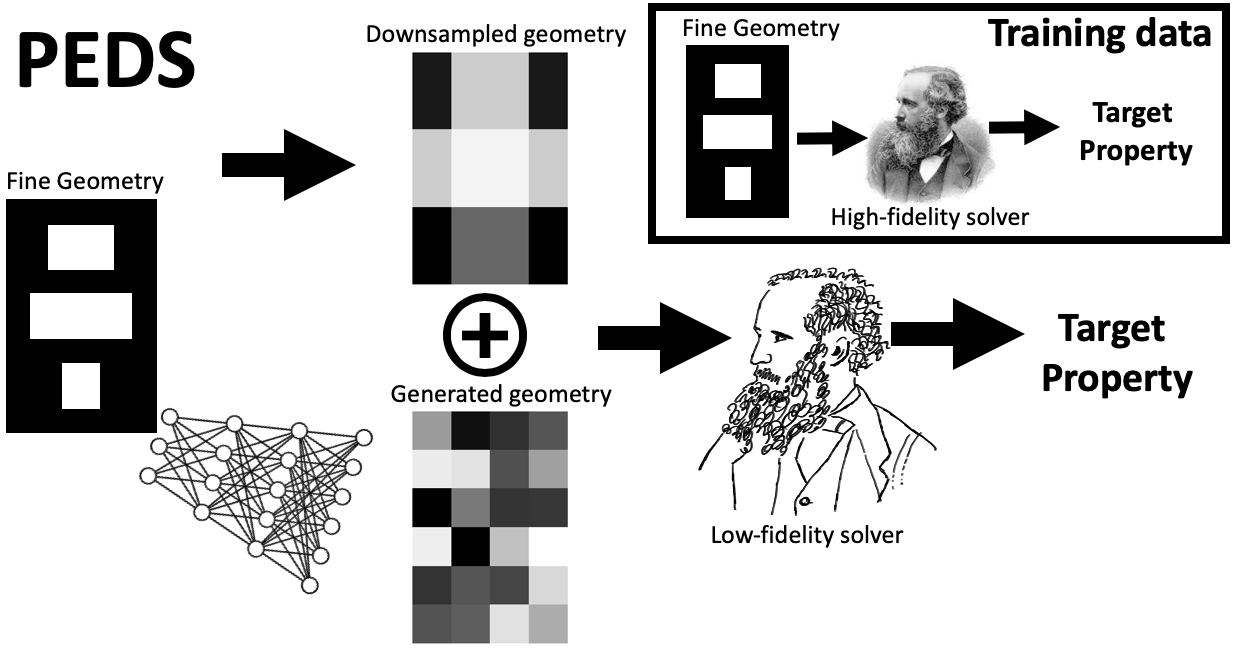}
    \caption{Diagram of PEDS: (Main) From the geometry parameterization, the surrogate generates a low-fidelity structure that is combined with a  downsampled geometry (e.g. downsampled by pixel averaging) to be fed into a low-fidelity solver (symbolized by a cartoon picture of James Clerk Maxwell). (Inset) The training data is generated by solving more costly simulations directly on a high-fidelity solver (symbolized by a photograph of James Clerk Maxwell).}
    \label{fig:PEDS_diagram}
\end{figure}
Before delving into implementation details and results, we present the core principles of PEDS which are common between all surrogates.

\begin{enumerate}
    \item Given the parameters $p$ of the geometry,
a deep generative NN model yields a grid of pixels describing a 
low-fidelity geometry.  We call this function $\mathrm{generator}_\mathrm{NN}(p)$.

    \item We also compute a low-fidelity downsampling (e.g. via sub-pixel averaging~\cite{oskooi2009accurate}) of the geometry, denoted $\mathrm{downsample}(p)$; other prior knowledge could also be incorporated here as well.
    
    \item We define $G$ as a weighted combination $G(p) = w\cdot \mathrm{generator}_\mathrm{NN}(p) + (1-w)\cdot \mathrm{downsample}(p)$, with a weight $w\in[0,1]$ (independent of $p$) that is another learned parameter.
    
    \item If there are any additional constraints/symmetries that the physical problem imposes on the geometry, they can be applied as projections $P[G]$.  For example, mirror symmetry could be enforced by averaging $G$ with its mirror image.
    
    \item Finally, given the low-fidelity geometry $P[G(p)]$, we evaluate the low-fidelity solver $f^\mathrm{lf}$ to obtain the property of interest:  $\tilde{f}(p) = f^\mathrm{lf}(P[G(p)])$.
    
\end{enumerate}
In summary, the PEDS model $\tilde{f}(p)$ is
\begin{equation}
\tilde{f}(p) = f^\mathrm{lf}\left(P\left[w\cdot\mathrm{generator}_\mathrm{NN}(p) + (1-w)\cdot \mathrm{downsample}(p)\right]\right) \, .
    \label{eq:model}
\end{equation}
For technical and implementation  details of the PEDS framework, see Methods section and SI.

\label{sec:model}

\subsection{Evaluation of PEDS across disparate exemplar surrogates}
In this work, we illustrate PEDS with three well-known PDEs, as shown in Extended Data Table~\ref{tab:fourier}, which are implicated in wide varieties of important applications. First, we study the linear diffusion equation (Figure 2, left and middle), which has applications in materials science, information theory, biophysics and probability, among others. In particular, we train a surrogate model for the thermal flux, which is a useful design property for thermoelectrics. Second, we build a surrogate model for the nonlinear reaction-diffusion equation. This PDE is used in chemistry and its surrogates can influence the design of chemical reactors (Figure 2, left and right). Third, we model the complex transmission of Maxwell's equations through a parameterized structure, which is typically used in the design of optical metamaterials~\cite{pestourie2020active, pestourie2018inverse, pestourie2020assume}. See Methods section for details.

\begin{figure}[ht!]
    \centering
    \includegraphics[width=\textwidth]{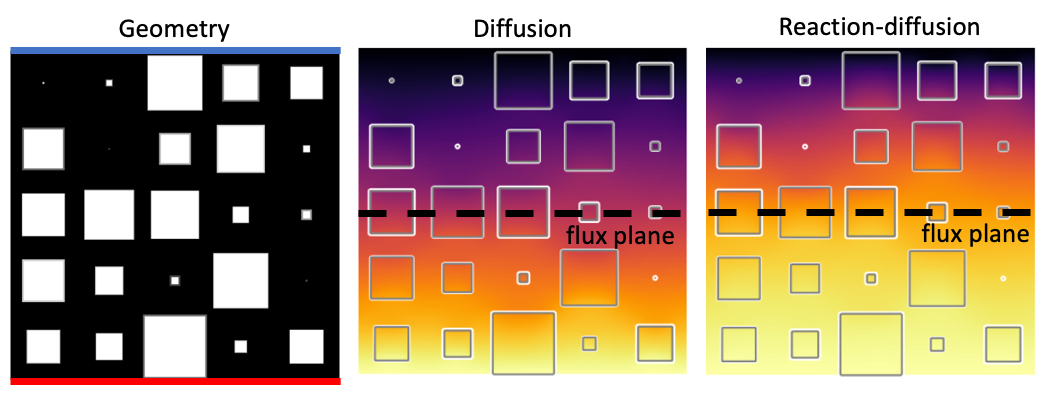}
    \caption{(Left) Geometry with 5 by 5 air holes with varying widths. There are Dirichlet boundary conditions on top (blue line) forcing the temperature to 0 and at the bottom (red line) forcing to 1, and periodic boundary conditions on the sides. (Middle and Right) Temperature field for the diffusion equation and the reaction diffusion equation, respectively. The orange dotted line is where the flux is evaluated to compute $\kappa$.}
    \label{fig:fffigure}
\end{figure}

\subsection{Overall benefits of PEDS} Most importantly, in a low-data regime ($\approx 10^3$ data points for 10 to 25 input parameters), we found for several very different PDEs that PEDS consistently increased the accuracy by up to $3\times$ and reduced the training data needed by at least an order of magnitude.
All PEDS surrogates reduce the need for training data by a factor of $>100$ to attain an error level of 5\% comparable to uncertainties in experiments (Extended data Table~\ref{tab:nnonlyresult}, Fig.~\ref{fig:resultfigure}), which is sufficient for design purposes. In the case of Fourier($16$) and Fourier($25$), the mixing weight  $w$ of the neural generated structures is around $0.1$, whereas for Fisher($16$) and Fisher($25$), the mixing weight $w$ is around $0.45$. Since the low-fidelity solver is more inaccurate for the nonlinear reaction--diffusion equation where the linear relaxation results in errors $>0.35\%$, the neural generator has approximately a $5\times$ larger weight, indicating it has the stronger impact of including the nonlinear effects in PEDS. We report the exact optimal combining weights in (SI, Table 1) for Fourier($16$), Fourier($25$), Fisher($16$), and Fisher($25$).
Performance in a low-data regime are summarized in Extended data Table~\ref{tab:nnonlyresult} for accuracy improvement, computed as the fractional error (FE) on a test set (SI, fractional error). For Fourier($16$), Fourier($25$), Fisher($16$), Fisher($25$), and Maxwell($10$), the error of PEDS goes down to typical levels of experimental uncertainties of 3.7\%, 3.8\%, 4.5\%, 5.5\%, and 19\% respectively.

We compared Fourier($16$), Fourier($25$), Fisher($16$), Fisher($25$), and Maxwell($10$) against a NN-only baseline, which consists of an ensemble of neural networks with the same number of parameters as PEDS generators with an additional fully connected layer to replace PEDS low-fidelity solver layer (Extended data Table~\ref{tab:nnonlyresult}). With 1000 training points, PEDS is an improvement compared to the neural network baseline of up to 3$\times$ (Extended data Table~\ref{tab:nnonlyresult}, PEDS ($\approx 10^3$) and NN-only ($\approx 10^3$)). Furthermore, the neural network baseline still cannot reach the reported PEDS accuracies when given an order of magnitude more data, which means that PEDS saves at least an order of magnitude in data (Extended data Table~\ref{tab:nnonlyresult}, NN-only ($\approx 10^4$)). Except Maxwell(10), the NN-only baselines cannot reach PEDS error with two orders of magnitude more data (Extended data Table~\ref{tab:nnonlyresult}, NN-only ($\approx 10^5$)). In particular for Fourier surrogates, going from $10^4$ to $10^5$ points reduces the error by less that $0.1\%$. Except Maxwell(10), which is further discussed in Section~\ref{sec:AL}, PEDS achieves error of $5\%$ in low-data regime (1000 training points), and reduces the data need by a factor of at least 100. We studied the general robustness of PEDS---the generalization error---with random split and stratified split (with respect to the magnitude of the target property) of the test set against the baselines of NN-only and prediction by the mean (SI, General robustness). We report that PEDS' error is 5x more robust for random splits in the test set. Also,  PEDS consistently shows significant improvement compared to the baselines for either random or stratified test set splits. 

We further compared PEDS to a low-fidelity solver baseline, which uses the low-fidelity solver with $\mathrm{downsample}(p)$ as input, without mixing with the low-fidelity geometry generated by the neural network (Extended data Table~\ref{tab:lowfidresult}).  PEDS also boosts the accuracy of the low-fidelity solver by $3.6\times$, $2.2\times$, $8.5\times$, $6.7\times$, and $6.5\times$, respectively (Extended data Table~\ref{tab:lowfidresult}, Improvement). For the reaction--diffusion equation, the low-fidelity solver has a coarser resolution and a linear approximation of the physics (neglecting the nonlinear term of reaction--diffusion equation), but the neural network generator captures the necessary nonlinearity to get improvement $> 5\times$ (Extended data Table~\ref{tab:lowfidresult}, Improvement).  The speedups vary between two and four orders of magnitude (Extended data Table~\ref{tab:lowfidresult}, Speedup). For Maxwell($10$), using a coarser low-fidelity solver generally gains two orders of magnitude in 2D, which should translate into a four orders of magnitude speedup for three-dimensional problems. We see the biggest speedups when the low-fidelity solver is not only coarser than the high-fidelity solver, but also when it is a linear relaxation of the physics (reaction--diffusion equation). In that case, the speedup is four orders of magnitudes.

\subsection{Detailed analysis of Maxwell(10) case study}\label{sec:AL}
In previous section, we showed the general performance of PEDS in the low-data regime. For Maxwell($10$), where the low-accuracy solver has a very large error ($>100\%$), we study the training curve asymptotically and when combining with AL~\cite{pestourie2020active}. In contrast to the previous section, where we performed static training that takes a training set sampled at random, here we discuss results from AL experiments by dynamic Bayesian training, where the training set is iteratively expanded using an AL algorithm~\cite{pestourie2020active}. Essentially, AL attempts to sample training points where the model uncertainty is highest, thereby reducing the number of costly point acquisitions by querying the high-fidelity solver. 
Our previous work showed an order of magnitude improvement in terms of data efficiency by using AL, when compared to a black-box NN~\cite{pestourie2020active}. Consistently, in this study, we also report substantial improvements from active learning for PEDS.

The active-learning algorithm iteratively builds a training set by filtering randomly generated points with respect to a trained measure of uncertainty~\cite{pestourie2020active}. The hyperparameters of this algorithm are (i) $n_\mathrm{init}$, which is the number of points the surrogate models are initially trained with; (ii) $T$, the number of exploration iteration; (iii) $M$ and $K$, which are such that $M\times K$ points are randomly generated at each iteration and only $K$ points with highest uncertainty $\sigma(p)$ are explored (SI, Active learning implementation details). We run the expensive high-fidelity solver to get the PDE solutions of the explored points. 
We have trained surrogates as well as an \emph{ensemble} of 5 independent surrogates. We found that models optimizing the negative log-likelihood perform similarly to models optimizing the mean squared error in the case static training. This is not surprising, because the mean squared error is part of the negative log-likelihood objective. Furthermore, the uncertainty model can be used as an approximation of the model error at evaluation time.
\label{sec:accuracy}

\begin{figure}[ht!]
    \centering
    \includegraphics[width=\textwidth]{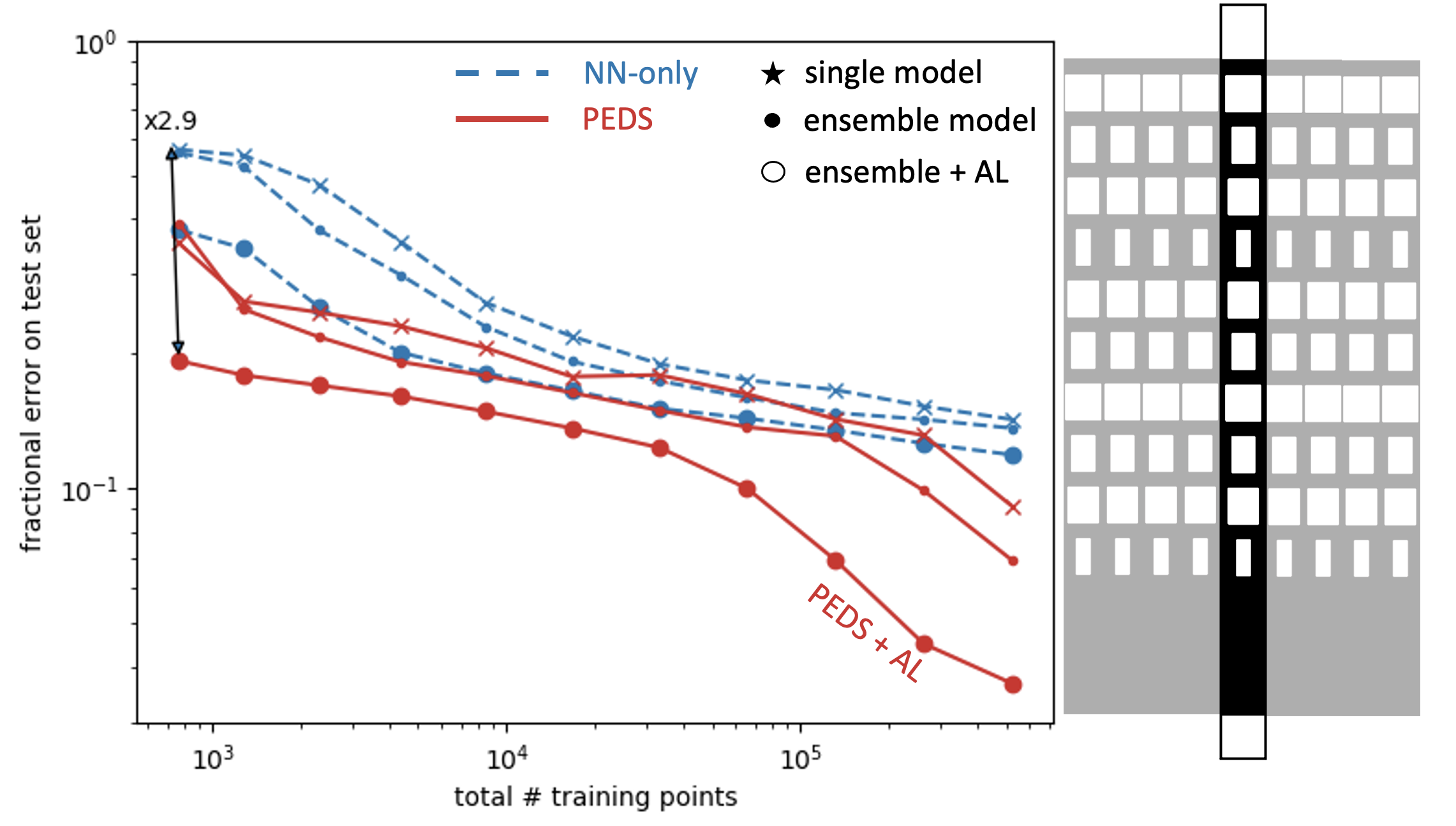}
    \caption{(Left) Fractional error (FE) on the test set: PEDS outperforms the other baseline models significantly when combined with active learning (AL). 
    (Right) Geometry of the unit cell of the surrogate model. Each of the 10 air holes have independent widths, the simulation is performed with periodic boundary conditions on the long sides, the incident light comes from the bottom and the complex transmission is measured at the top of the geometry.}
    \label{fig:resultfigure}
\end{figure}

We compared PEDS to a NN-only baseline 
using the fractional error as an evaluation metric~(SI, Implementation details of PEDS and baselines).
In Fig.~\ref{fig:resultfigure}, we show that PEDS clearly outperforms all other models when combined with active learning. In low-data regime, it is $2.9\times$ more accurate than the baseline. Asymptotically, in high-data regime, it converges to the true value with a power law exponent $5\times$ better, with a slope of -0.5, in contrast to -0.1, for the baseline on the loglog plot. 

From a data-efficiency perspective, the PEDS+AL solver achieves 20\% error on the test set, while using only about $5\%$ of the training data needed to train the NN-only baseline, and $12.5\%$ of the training data needed to train the NN-only baseline with AL (Fig.~\ref{fig:resultfigure}).
Only PEDS+AL reaches a low $3.5$\% error with a training data size of $\approx500k$ (Fig.~\ref{fig:resultfigure}). However, if we extrapolate the other curves in Fig.~\ref{fig:resultfigure}, it is clear that they would require at \emph{least} two orders of magnitude more data to achieve similar low error. This completes the claim that PEDS saves at least two orders of magnitude in training data to achieve and error comparable to fabrication uncertainty.

Evaluating the baseline (with an ensemble of neural networks) takes 500~$\mu s$, while PEDS evaluates in $5$~ms, which is about a ten times slower. However the high-fidelity solver is about a hundred times slower, evaluating at $\approx1$~s. In order to simulate the data set quickly, and without loss of generality, we showed results for PEDS in 2D (Fig.~\ref{fig:resultfigure}~(right). As PEDS is already faster than the high-fidelity model by two orders of magnitude, this difference will be even starker for 3D simulations. The simulation of the equivalent structure in 3D evaluates in about $100$~ms with the low-fidelity model, and in $2462$~s with the high-fidelity model. In this occurrence, PEDS would represent a speed-up by at least four orders of magnitude.  
For ablation experiments, an analysis of generated geometries, and a comparison to mainstream surrogate models, see SI.

\section{Discussion}
\label{sec:discussion}

The significance of the PEDS approach is that it can easily be applied to a wide variety of physical systems.  It is common across many disciplines to have models at varying levels of fidelity, whether they simply differ in spatial resolution (as in Fourier($16$), Fourier($25$), and Maxwell($10$)) or in the types of physical processes they incorporate (as in Fisher($16$) and Fisher($25$)). 
For example, in fluid mechanics the low-fidelity model could be Stokes flow (neglecting inertia), while the high-fidelity model might be a full Navier--Stokes model (vastly more expensive to simulate)~\cite{ferziger2002computational}, with generator NN correcting for the deficiencies of the simpler model.  As another example, we are currently investigating a PEDS approach to construct a surrogate for complex Boltzmann-transport models~\cite{romano2021openbte} where the low-fidelity heat-transport equation can simply be a diffusion equation. \edit{The last layer, in principle, does not require a direct solver. It can also backpropagate from a nonlinear solver using the implicit function theorem. The only criteria needed from the low-fidelity solver are the inclusion criterion and the criterion that it evaluates significantly faster than the high-fidelity solver. We leave the evaluation of nonlinear solver in the last layer for future work.}

Knowledge of priors can also be introduced in the low-fidelity geometry that is mixed with the neural generator output. PEDS provides a data-driven strategy to connect a vast array of simplified physical models with the accuracy of brute-force numerical solvers, offering both more insight and more data efficiency than physics-independent black-box surrogates. The low-fidelity model in PEDS contains the physics and has the same generality as its class of low-fidelity model. The model is no longer a black box and the neural network generator can be inspected after training to ensure that the output makes intuitive physical sense (Analysis of generated geometries) or the generator can be sparsified~\cite{cranmer2020discovering, rackauckas2020universal}. \edit{For further intuition on why PEDS' accuracy is not limited by the low-fidelity solver's accuracy, see  SI on when/how the universal approximation theorem can be extended to PEDS.} In future work, we will consider how the regularization from the low-fidelity solver layer may be optimal, similar to the results of other models such as ridge regression~\cite{hoerl1970ridge}.

When compared to related works, PEDS should not be confused with physics-informed neural networks~(PINNs), which solve the full PDE (imposed pointwise throughout the domain) for the entire PDE solution (\emph{not} a surrogate for a finite set of outputs like the complex transmission or the thermal flux)~\cite{karniadakis2021physics, lu2021physics}, and which do not employ any pre-existing solver. Current PINNs tend to be slower than conventional high-fidelity PDE solvers (e.g. based on finite elements)~\cite{shin2020convergence}, but offer potentially greater flexibility. Universal ordinary differential equations (UODEs)~\cite{rackauckas2020universal} and machine-learning-accelerated~\cite{kochkov2021machine} methods also tackle a different problem from PEDS: they identify unknown dynamics in an ODE by replacing the unknown terms with neural networks trained on data. In contrast to DeepONet~\cite{lu2021learning, lu2022multifidelity} and Fourier neural operators~\cite{li2020fourier} which are solvers to PDES, PEDS is a surrogate model which predicts an output property given a parameterization. Consequently, PEDS works with experimental property data when the generative high-fidelity model is unknown. \edit{Note that using a DeepONet as a surrogate model reduces to our NN-only baseline. Since the output of a surrogate model is independent of spatio-temporal coordinates, the trunk net reduces to a constant.} In the future, PEDS will be studied in the context of model discovery.  Our approach has some similarities with input space mapping (SM)~\cite{koziel2008space}, especially neural SM~\cite{bakr2000neural} and coarse mesh/fine SM~\cite{feng2019coarse}, where the input of a fine solver is mapped into the input of a coarse solver. However SM uses the same parameterization for the fine solver and the coarse solver, rather than mapping to ``downsampled'' resolution, and does not mix the generated input with a downsampled guess adaptively. We show that PEDS substantially outperforms SM in the SI (SM baseline). Finally, in contrast to error-correction techniques at the output level of the surrogate~\cite{lu2020extraction, koziel2006space, levine2022framework, ren2022physics}, generative-model-only techniques~\cite{drygala2022generative, geneva2020multi}, and reduced-order models~\cite{hesthaven2022reduced}, PEDS includes the solver in an end-to-end fashion during the training process. In PEDS, the output of the low-fidelity solver layer is not further transformed, which preserves key properties of the low-fidelity solver such as conservation of energy or mass. Mappings between coarse and fine descriptions of a system is also leveraged in the renormalization group technique in physics~\cite{weinberg1995quantum}, but in the latter context this is accompanied by a change of scale---often to investigate self-similar phenomena---and not necessarily a change in the number of degrees of freedom. PEDS can hierarchically combine models with varying levels of complexity such as low- and mid-fidelity solvers, or linearized solver and coarse nonlinear solver. This could be implemented by learning mixing parameters for the different models and requiring them to be sparse with a L1 regularization to reduce the evaluation costs. In a similar way, PEDS can be used in a multiscale setting as in ~\citeasnoun{hou2017iteratively}.

In addition to applying the PEDS approach to additional physical systems, there are a number of other possible technical refinements.   For example, one could easily extend the PEDS NN to take an image of the high-fidelity-structure geometry rather than its parameterization, perhaps employing  convolutional neural networks to represent a translation-independent ``coarsification'' and/or a multiresolution architecture.    This type of surrogate could then be employed for topology optimization in which ``every pixel'' is a degree of freedom~\cite{molesky2018inverse}.  Another interesting direction might be to develop new low-fidelity
physics models that admit ultra-fast solvers but are too inaccurate to be used \emph{except} with PEDS;
for instance, mapping Maxwell's equations in 3D onto a simpler
(scalar-like) wave equation or mapping the materials into
objects that admit especially efficient solvers (such
as impedance surfaces~\cite{perez2018sideways} or compact objects for surface-integral equation methods~\cite{jin2015finite}).

\section*{Methods}
\edit{\paragraph{Inclusion criterion for PEDS} The existence criterion of an arbitrary accurate physics-enhanced deep surrogate is that the span of the high-fidelity property be included in the span of the property computed with the low-fidelity solver. This criterion should guide the choice of the low-fidelity solver for PEDS. If this condition is not met by the low-fidelity solver, then it is not expressive enough to capture the full range of output of the target property function.}

\paragraph{Optimization starting point} PEDS can be initialized very close to the low-fidelity solver by setting $w=0.05$. Using the low-fidelity model as an initial guess for PEDS training imposes a lower bound (= low-fidelity solver performance) on the performance of PEDS on the training set. Results show that indeed PEDS performance never degrades below the low-fidelity solver. In fact, we find that PEDS performs significantly better than the low-fidelity solver alone on a held-out test set. Most optimization algorithms will not converge in practice to a solution worse than the starting point. Some algorithms can guarantee a sufficient decrease condition like CCSA~\cite{svanberg2002class}, LBFGS~\cite{liu1989limited}, etc, to ensure that PEDS does not degrade the low-fidelity solver. The choice of physical inductive bias to add as constraints in the optimization has been determined by an ablation study (\secref{sec:AL}, Ablation Study).

\paragraph{Dataset acquisition } PEDS is a supervised model that is trained on a labeled dataset.  It is not necessary to know the generative high-fidelity model for PEDS. The data can be generated from real-world experiments. But in the case of our illustrative examples, we build the training set by querying the high-fidelity solver with parameterized geometries $S=\{ (p_i, t^{hf}_i) , i=1 ... N\}$, where $p_i$ are parameterized geometries in the training set and $t^{hf}_i=f^{hf}(p_i)$. The upfront cost of building the training dataset is the most time-consuming part of developing a supervised surrogate model $\tilde{f}(p)$. By building some approximate low-fidelity physics knowledge into the surrogate, we will show that PEDS greatly reduces the number $N$ of queries to expensive simulations. Note that the generation of the high-fidelity data is embarrassingly parallel when the high-fidelity model is known.

\paragraph{Training loss }A basic PEDS training strategy could simply minimize the mean squared error $\sum_{(p,t^\mathrm{hf})\in S}|\tilde{f}(p) - t^\mathrm{hf}|^2$ (for a training set $S$) with respect to the parameters of the NN and the weight~$w$. When the data may have outliers, we use a Huber loss~\cite{huber1992robust}.

\begin{equation}\label{eq:huber}
    L_\delta (a) = \begin{cases}
 \frac{1}{2}{a^2}                   & \text{for } |a| \le \delta, \\
 \delta \cdot \left(|a| - \frac{1}{2}\delta\right), & \text{otherwise.}
\end{cases}
\end{equation}

We also employ a  more complicated loss function that allows us to easily incorporate active-learning strategies~\cite{pestourie2020active}.  We optimize the Gaussian negative log-likelihood of a Bayesian model~\cite{lakshminarayanan2016simple}
\begin{equation}\label{eq:loglikelihood}
    -\sum_{(p_i, t^{hf}_i)\in S} \log{\mathrm{P}_\Theta(t^{hf}_i|p_i)} \propto \sum_{(p_i, t^{hf}_i)\in S} \left[ \log{\sigma(p_i)} + \frac{(t^{hf}_i-\tilde{f}(p_i))^2}{2 \sigma(p_i)^2} \right]
\end{equation}
where $\mathrm{P}_\Theta$ is a Gaussian likelihood defined by $\Theta$ which includes the parameters of the generator model parameters and the combination weight $w$, and the heteroskedastic ``standard deviation'' $\sigma(p) > 0$ is the output of another NN (trained along with our surrogate model). 

\paragraph{Ensemble model} We also train surrogates that are an \emph{ensemble} of 5 independent surrogates. The prediction of the ensemble is the average of the predictions of each individual model. A model error can be inferred at inference time from the diversity of predictions~\cite{lakshminarayanan2016simple}.

\paragraph{Stochastic gradient descent }In practice, rather than examining the entire training set $S$ at each training step, we follow the standard ``batch'' approach~\cite{goodfellow2016deep} of sampling a random subset of $S$ and minimizing the expected loss with the Adam stochastic gradient-descent algorithm~\cite{kingma2014adam} (via the Flux.jl~\cite{innes:2018} software in the Julia language).

\paragraph{Adjoint method} The low-fidelity solver is a layer of the PEDS model, which is trained end-to-end, so we must backpropagate its gradient $\nabla_g f^\mathrm{lf}$ with respect to the low-fidelity geometry input $g$ through the other layers to obtain the overall sensitivities of the loss function. This is accomplished efficiently using the known ``adjoint'' methods~\cite{molesky2018inverse}. Such methods  yield a vector-Jacobian product that is then automatically composed with the other layers using automatic differentiation~(AD) (via the Zygote.jl~\cite{innes2018don} software). 

In particular, the low-fidelity solver layer is differentiable because each pixel of the low-fidelity geometry is assigned to a sub-pixel average of the infinite-resolution structure, which increases accuracy~\cite{oskooi2009accurate} and makes $\mathrm{downsample}(p)$ piecewise differentiable. In the same way, $\mathrm{hf}(p)$ is differentiable for the high-fidelity geometry.

\paragraph{PEDS for diffusion equation} Our first two surrogate models are for the diffusion equation from Extended data Table~\ref{tab:fourier}. They are called Fourier($16$) and Fourier($25$), and they predict the thermal flux $\kappa(p)$ from the diffusion equation for 16 and 25 input variables, respectively. As showed in Fig.~\ref{fig:fffigure}~(left),  the 2D nanostructured material defines the coefficient matrix $D(p)$ where the parameter vector $p$ contains the 25 (resp. 16) independent side lengths of a five by five (resp. four by four) grid of air holes etched in the medium. The thermal conductivity coefficients in $D$ are set to 1 in the medium and 0.1 in the holes. The boundary conditions are periodic in $x$-direction and Dirichlet boundary conditions in the $y$ direction, fixing the temperature to $1$ at the bottom and to $0$ at the top, as illustrated by thick red and blue lines in Fig.~\ref{fig:fffigure}~(left). The Dirichlet boundary conditions are equivalent to the source term $\textbf{s}_0$ in Extended data Table~\ref{tab:fourier}.
Both the high-fidelity and the low-fidelity solvers employ a finite-difference solver that represents the geometry by a grid of discretized thermal conductivity. Sub-pixel averaging is employed at the boundary between the holes and the medium. For both Fourier($16$) and Fourier($25$), the high-fidelity solver has a resolution of 100. The low-fidelity solver has a resolution of 4 or 5, which corresponds to a single pixel per hole position. Each high-fidelity data point acquisition requires $\approx 35$~ms, and each low-fidelity data point acquisition requires $\approx 65~\mu$m and $\approx 75~\mu$m, respectively, which represents a speed-up of $\approx 500\times$ (Extended data Table~\ref{tab:lowfidresult}, Speedup). We compute the low-fidelity solver baseline error, by computing the solution with the low-fidelity solver and the geometry $\mathrm{downsample}(p)$, where $p$ is the geometry parameterization (i.e. without mixing with a neural generator output). Despite the much lower resolution, the low fidelity solvers have a fairly low error of 13.5\% and 8.5\%, respectively. This good performance of an averaged structure comes from the fact that the diffusion equation is a smoothing equation. Nonetheless, such errors would still be dominant compared to typical experimental uncertainties of $\approx$5\%.
Fourier($16$) and Fourier($25$) were trained to predict the flux through a plane as in Fig.~\ref{fig:fffigure}~(middle) by minimizing Huber loss in Eq.~\ref{eq:huber} with $\delta=10^{-3}$ to lower the sensitivity to outliers.

\paragraph{PEDS for reaction--diffusion equation} Our next two surrogate models solve the reaction--diffusion equation from Extended data Table~\ref{tab:fourier}, and are called Fisher($16$) and Fisher($25$). They predict the flux $\kappa(p)$ through the same geometry as Fourier($16$) and Fourier($25$), respectively. As can be seen in Extended data Table~\ref{tab:fourier} the reaction--diffusion equation has an additional nonlinear term $k \textbf{u}(1-\textbf{u})$ compared to the diffusion equation. $k$ is a coefficient that controls the amount of nonlinearity in the PDE. In Fig.~\ref{fig:fffigure}~(middle and right), we see how much the nonlinearity impacts the PDE solution. The high-fidelity nonlinear solver is using finite difference and Newton's method in conjunction with a continuation method that increases $\kappa$ from 0.1 to 10 in 5 multiplicative steps. The low-fidelity solvers of Fisher($16$) and Fisher($25$) are identical to that of Fourier($16$) and Fourier($25$), respectively. Importantly, the low-fidelity solver not only has a coarse resolution, but also uses an approximate physics that neglects the nonlinear term from the reaction--diffusion equation. Each high-fidelity data point requires $\approx700$~ms that is around $10^4\times$ slower than the low-fidelity solver (Extended data Table~\ref{tab:lowfidresult}, Speedup). The low-fidelity solvers have error of 38.1\% and 36.7\% respectively. Fisher($16$) and Fisher($25$)  were trained to predict the flux through a plane as in Huber loss in Eq.~\ref{eq:huber} with $\delta=10^{-3}$ to lower the sensitivity to outliers.

\paragraph{PEDS for Maxwell's equations} Similarly to~\citeasnoun{pestourie2020active}, our third surrogate model Maxwell($10$) predicts the complex transmission $t^{hf}(p)$ of a 2D ``meta-atom'' unit cell with a parameterized geometry $p$, which consists of ten layers of air holes with independent widths etched in a substrate (of dielectric constant $\varepsilon = 2.1$ corresponding to silica), with periodic boundary conditions in $x$ and outgoing radiation boundary conditions in the $y$ direction and an incoming normal-incident planewave from below, as shown in Fig.~\ref{fig:resultfigure}~(right). 
In terms of the vacuum wavelength $\lambda$ of the incident wave (for the largest $\lambda$ considered below), the period in $x$ is $0.95\lambda$ and the total thickness is $11\lambda$ (with hole heights of 0.75$\lambda$ and interstices of 0.35$\lambda$); the fact that the structure is several wavelengths in diameter causes the transmission $t^{hf}(p)$ to be a complicated oscillatory function that makes the surrogate training challenging~\cite{pestourie2020active}.  A ``metasurface'' consists of a collection of many of these meta-atoms, designed to perform some optical function such as focusing~\cite{li2021inverse}.  The full solution for a metasurface can be approximated in terms of the transmissions of individual periodic `unit cells via a local periodic approximation~\cite{pestourie2018inverse, pestourie2020assume}. A schematic unit cell with 3~holes is showed in Fig.~\ref{fig:PEDS_diagram}, and an example 10-hole structure from the training set is shown in Fig.~\ref{fig:resultfigure}~(right).

Both the high-fidelity and low-fidelity solvers
for Maxwell($10$) employ finite-difference frequency-domain (FDFD) discretizations of Maxwell's equations~\cite{champagne2001fdfd}, using perfectly matched layers (PMLs)~\cite{sacks1995perfectly} to implement outgoing boundary conditions.  Similarly to the solvers of the two previous equations,  FDFD represents the geometry by a grid of discretized $\varepsilon$ ``pixels,'' which is a function of the parameters (hole widths) $p$, $\mathrm{hf}(p)$, and $\mathrm{downsample}(p)$ for the high-fidelity solver and the baseline coarse solver, respectively. An FDFD resolution of 40 pixels per wavelength is used as our high-fidelity solver.
This resolution is typical for high-fidelity solvers in electromagnetism, because it is comparable to the manufacturing accuracy in nanophotonics and hence suffices for practical metalens design~\cite{li2021inverse, bayati2021inverse} within fabrication uncertainty. (Sharp/narrowband resonances can shift if one refines the resolution further, but the positions and the bandwidths of the resonances are accurate to within a few percent.) Each high-fidelity-solver data point required $\approx 1$~s (on a 3.5 GHz 6-Core Intel Xeon E5); an analogous simulation in 3D takes several hours.  Our PEDS surrogate uses an FDFD solver at a coarser resolution of 10 pixels per wavelength, which is about $100\times$ faster in 2D and $> 10^4\times$ faster in 3D, but has much worse accuracy. 
It differs from the high-fidelity solver's transmission by $124$\% on our test set, which is significantly more than the four other surrogates presented in this article. Maxwell($10$) model was trained to predict the complex transmission for 3 frequencies by minimizing the negative Gaussian likelihood loss function to enable comparison with and without using AL~\cite{pestourie2020active}. The input of the model $p$ is the concatenation of the 10 widths and the one-hot encoding of the frequency. 

\section*{Data Availability Statement}

 Datasets to reproduce the findings for the five surrogate models with about 1000 data points are available on GitHub\cite{rpestourie_payel79_2023}. The full dataset and the raw data to create the figures of the main text can be available from the corresponding authors upon reasonable request.

\section*{Code Availability statement}
The code used for these findings can be found on GitHub\cite{rpestourie_payel79_2023}. 

\section*{Acknowledgements}

R.P. was supported by the U.S. Army Research Office through the Institute for Soldier Nanotechnologies (Award No. W911NF-18-2-0048) and the MIT-IBM Watson AI Laboratory. The authors thank Meredith Dost for her suggestions in proof reading.

\section*{Author contributions}
R.P., Y.M., C.R., P.D., and S.G.J. designed the study, contributed to the machine-learning approach, and analyzed results; R.P. led the code development, software implementation, and numerical experiments; R.P. and S.G.J. were responsible for the physical ideas and interpretation. All authors contributed to the algorithmic ideas and writing.

\section*{Competing interests}
The authors declare no competing financial or non-financial interests.

\section*{Tables}
\begin{table}[h!]
\centering
\begin{tabular}{lll}
\hline
Equation name             & Equation formula & Model(\textit{input dimension})\\
\hline
Diffusion          & $\nabla\cdot D\nabla \textbf{u}= \textbf{s}_0$ & Fourier($d$)\\ 
Reaction-diffusion &  $\nabla\cdot D\nabla \textbf{u}= -k\textbf{u}(1-\textbf{u})+\textbf{s}_0$ & Fisher($d$)\\ 
2D Maxwell (Helmholtz)          & $\nabla^2\textbf{u}+\omega^2\varepsilon\textbf{u}=\textbf{s}_1$ & Maxwell($d$)\\
\hline
\end{tabular}
\setcounter{table}{0}
\renewcommand{\tablename}{Extended data Table}
\caption{Governing equations of the surrogate models for our example problems. $d$ is the input dimension, i.e. the number of input variables in the surrogate model, which ranges from $10$ to $25$.}
\label{tab:fourier}
\end{table}

\begin{table}[h!]
\begin{tabular}{lllll}
\hline
Model(\textit{input dim})  & PEDS ($\approx 10^3$) & NN-only ($\approx 10^3$) & NN-only ($\approx 10^4$) & NN-only ($\approx 10^5$) \\ 
\hline
Fourier(16) & 3.7\% & 5.1\% & 4.8\% & 4.8\% \\
Fourier(25) & 3.8\% & 4.7\% & 4.4\% & 4.4\% \\
Fisher(16)  & 4.5\% & 10.1\% & 9.9\% & 9.5\% \\
Fisher(25)  & 5.5\% & 14.4\% & 14.0\% & 12.7\% \\
Maxwell(10)  & 19\% (AL) & 56\% & 19\% & 15\% \\
\hline
\end{tabular}
\renewcommand{\tablename}{Extended data Table}
\caption{PEDS error versus NN-only baselines' errors (fractional error on the test set). We report the orders of magnitude of training points in parenthesis. With more than an order of magnitude extra data, NN-only baseline still has much higher error than PEDS.  Except Maxwell(10), all baselines still cannot achieve PEDS error with two orders of magnitude extra data. The improvement when going from $10^4$ to $10^5$ points with Fourier surrogates are smaller than $0.1\%$. In the Maxwell case, we show in section 3.3 that it is crucial to include active learning (AL) in addition to PEDS.}
\label{tab:nnonlyresult}
\end{table}

\begin{table}[h!]
\begin{tabular}{llllll}
\hline
Model(\textit{input dim}) & PEDS error ($\approx 10^3$) & Low-fidelity error & Improvement & Speedup \\
\hline
Fourier(16) & 3.7\% & 13.5\% & $3.6\times$ & 500$\times$ 
\\
Fourier(25) & 3.8\% & 8.5\% & $2.2\times$ & 500$\times$ 
\\
Fisher(16) & 4.5\% & 38.1\% & $8.5\times$ & $10^4\times$ 
\\   
Fisher(25) & 5.5\% & 36.7\% & $6.7\times$ & $10^4\times$ 
\\    
Maxwell(10) & 19\% (AL) & 124\%  & $6.5\times$ & $10^2\times$ / $10^4\times$\\
\hline
\end{tabular}
\renewcommand{\tablename}{Extended data Table}
\caption{With $\approx 10^3$ training points, PEDS consistently improves error (fractional error on the test set) by 2--8$\times$ compared to the low-fidelity solver. ``Improvement'' is the reduction in error by PEDS compared to the low-fidelity.  Speedups are shown for 2D simulations, and speedup for 3D simulations is also reported for Maxwell($10$)}
\label{tab:lowfidresult}
\end{table}

\clearpage
\newpage
\section*{Figure Captions}

\paragraph{Figure 1} Diagram of PEDS: (Main) From the geometry parameterization, the surrogate generates a low-fidelity structure that is combined with a  downsampled geometry (e.g. downsampled by pixel averaging) to be fed into a low-fidelity solver (symbolized by a cartoon picture of James Clerk Maxwell). (Inset) The training data is generated by solving more costly simulations directly on a high-fidelity solver (symbolized by a photograph of James Clerk Maxwell).

\paragraph{Figure 2} (Left) Geometry with 5 by 5 air holes with varying widths. There are Dirichlet boundary conditions on top (blue line) forcing the temperature to 0 and at the bottom (red line) forcing to 1, and periodic boundary conditions on the sides. (Middle and Right) Temperature field for the diffusion equation and the reaction diffusion equation, respectively. The orange dotted line is where the flux is evaluated to compute $\kappa$.

\paragraph{Figure 3} (Left) Fractional error (FE) on the test set: PEDS outperforms the other baseline models significantly when combined with active learning (AL). (Right) Geometry of the unit cell of the surrogate model. Each of the 10 air holes have independent widths, the simulation is performed with periodic boundary conditions on the long sides, the incident light comes from the bottom and the complex transmission is measured at the top of the geometry.

\clearpage
\newpage
\section*{References}
\bibliographystyle{naturemag}
\bibliography{refs}

\begin{thebibliography}{10}
\expandafter\ifx\csname url\endcsname\relax
  \def\url#1{\texttt{#1}}\fi
\expandafter\ifx\csname urlprefix\endcsname\relax\def\urlprefix{URL }\fi
\providecommand{\bibinfo}[2]{#2}
\providecommand{\eprint}[2][]{\url{#2}}

\bibitem{baker2019workshop}
\bibinfo{author}{Baker, N.} \emph{et~al.}
\newblock \bibinfo{title}{Workshop report on basic research needs for
  scientific machine learning: Core technologies for artificial intelligence}.
\newblock \bibinfo{type}{Tech. Rep.}, \bibinfo{institution}{USDOE Office of
  Science (SC), Washington, DC (United States)} (\bibinfo{year}{2019}).

\bibitem{benner2015survey}
\bibinfo{author}{Benner, P.}, \bibinfo{author}{Gugercin, S.} \&
  \bibinfo{author}{Willcox, K.}
\newblock \bibinfo{title}{A survey of projection-based model reduction methods
  for parametric dynamical systems}.
\newblock \emph{\bibinfo{journal}{SIAM Review}} \textbf{\bibinfo{volume}{57}},
  \bibinfo{pages}{483--531} (\bibinfo{year}{2015}).

\bibitem{willard2020integrating}
\bibinfo{author}{Willard, J.}, \bibinfo{author}{Jia, X.}, \bibinfo{author}{Xu,
  S.}, \bibinfo{author}{Steinbach, M.} \& \bibinfo{author}{Kumar, V.}
\newblock \bibinfo{title}{Integrating physics-based modeling with machine
  learning: A survey}.
\newblock \emph{\bibinfo{journal}{arXiv preprint arXiv:2003.04919}}
  (\bibinfo{year}{2020}).

\bibitem{hoffmann2019machine}
\bibinfo{author}{Hoffmann, J.} \emph{et~al.}
\newblock \bibinfo{title}{Machine learning in a data-limited regime: Augmenting
  experiments with synthetic data uncovers order in crumpled sheets}.
\newblock \emph{\bibinfo{journal}{Science Advances}}
  \textbf{\bibinfo{volume}{5}}, \bibinfo{pages}{eaau6792}
  (\bibinfo{year}{2019}).

\bibitem{pant2021deep}
\bibinfo{author}{Pant, P.}, \bibinfo{author}{Doshi, R.}, \bibinfo{author}{Bahl,
  P.} \& \bibinfo{author}{Barati~Farimani, A.}
\newblock \bibinfo{title}{Deep learning for reduced order modelling and
  efficient temporal evolution of fluid simulations}.
\newblock \emph{\bibinfo{journal}{Physics of Fluids}}
  \textbf{\bibinfo{volume}{33}}, \bibinfo{pages}{107101}
  (\bibinfo{year}{2021}).

\bibitem{pestourie2018inverse}
\bibinfo{author}{Pestourie, R.} \emph{et~al.}
\newblock \bibinfo{title}{Inverse design of large-area metasurfaces}.
\newblock \emph{\bibinfo{journal}{Optics Express}}
  \textbf{\bibinfo{volume}{26}}, \bibinfo{pages}{33732--33747}
  (\bibinfo{year}{2018}).

\bibitem{boyd2007chebyshev}
\bibinfo{author}{Boyd, J.~P.}
\newblock \emph{\bibinfo{title}{Chebyshev and Fourier Spectral Methods, 2nd
  edn}} (\bibinfo{publisher}{Dover Publications, Inc., Minoela},
  \bibinfo{year}{2001}).

\bibitem{pestourie2020active}
\bibinfo{author}{Pestourie, R.}, \bibinfo{author}{Mroueh, Y.},
  \bibinfo{author}{Nguyen, T.~V.}, \bibinfo{author}{Das, P.} \&
  \bibinfo{author}{Johnson, S.~G.}
\newblock \bibinfo{title}{Active learning of deep surrogates for pdes:
  application to metasurface design}.
\newblock \emph{\bibinfo{journal}{npj Computational Materials}}
  \textbf{\bibinfo{volume}{6}}, \bibinfo{pages}{1--7} (\bibinfo{year}{2020}).

\bibitem{lu2022multifidelity}
\bibinfo{author}{Lu, L.}, \bibinfo{author}{Pestourie, R.},
  \bibinfo{author}{Johnson, S.~G.} \& \bibinfo{author}{Romano, G.}
\newblock \bibinfo{title}{Multifidelity deep neural operators for efficient
  learning of partial differential equations with application to fast inverse
  design of nanoscale heat transport}.
\newblock \emph{\bibinfo{journal}{arXiv preprint arXiv:2204.06684}}
  (\bibinfo{year}{2022}).

\bibitem{pestourie2020assume}
\bibinfo{author}{Pestourie, R.}
\newblock \emph{\bibinfo{title}{Assume Your Neighbor is Your Equal: Inverse
  Design in Nanophotonics}}.
\newblock Ph.D. thesis, \bibinfo{school}{Harvard University}
  (\bibinfo{year}{2020}).

\bibitem{bayati2021inverse}
\bibinfo{author}{Bayati, E.} \emph{et~al.}
\newblock \bibinfo{title}{Inverse designed extended depth of focus meta-optics
  for broadband imaging in the visible}.
\newblock \emph{\bibinfo{journal}{arXiv preprint arXiv:2105.00160}}
  (\bibinfo{year}{2021}).

\bibitem{li2021inverse}
\bibinfo{author}{Li, Z.} \emph{et~al.}
\newblock \bibinfo{title}{Inverse design enables large-scale high-performance
  meta-optics reshaping virtual reality}.
\newblock \emph{\bibinfo{journal}{arXiv preprint arXiv:2104.09702}}
  (\bibinfo{year}{2021}).

\bibitem{potton2004reciprocity}
\bibinfo{author}{Potton, R.~J.}
\newblock \bibinfo{title}{Reciprocity in optics}.
\newblock \emph{\bibinfo{journal}{Reports on Progress in Physics}}
  \textbf{\bibinfo{volume}{67}}, \bibinfo{pages}{717} (\bibinfo{year}{2004}).

\bibitem{li2021trustworthy}
\bibinfo{author}{Li, B.} \emph{et~al.}
\newblock \bibinfo{title}{Trustworthy ai: From principles to practices}.
\newblock \emph{\bibinfo{journal}{arXiv preprint arXiv:2110.01167}}
  (\bibinfo{year}{2021}).

\bibitem{oskooi2009accurate}
\bibinfo{author}{Oskooi, A.~F.}, \bibinfo{author}{Kottke, C.} \&
  \bibinfo{author}{Johnson, S.~G.}
\newblock \bibinfo{title}{Accurate finite-difference time-domain simulation of
  anisotropic media by subpixel smoothing}.
\newblock \emph{\bibinfo{journal}{Optics letters}}
  \textbf{\bibinfo{volume}{34}}, \bibinfo{pages}{2778--2780}
  (\bibinfo{year}{2009}).

\bibitem{ferziger2002computational}
\bibinfo{author}{Ferziger, J.~H.}, \bibinfo{author}{Peri{\'c}, M.} \&
  \bibinfo{author}{Street, R.~L.}
\newblock \emph{\bibinfo{title}{Computational methods for fluid dynamics}},
  vol.~\bibinfo{volume}{3} (\bibinfo{publisher}{Springer},
  \bibinfo{year}{2002}).

\bibitem{romano2021openbte}
\bibinfo{author}{Romano, G.}
\newblock \bibinfo{title}{Openbte: a solver forab-initio phonon transport in
  multidimensional structures}.
\newblock \emph{\bibinfo{journal}{arXiv preprint arXiv:2106.02764}}
  (\bibinfo{year}{2021}).
\newblock \urlprefix\url{https://arxiv.org/abs/2106.02764}.

\bibitem{cranmer2020discovering}
\bibinfo{author}{Cranmer, M.} \emph{et~al.}
\newblock \bibinfo{title}{Discovering symbolic models from deep learning with
  inductive biases}.
\newblock \emph{\bibinfo{journal}{Advances in Neural Information Processing
  Systems}} \textbf{\bibinfo{volume}{33}}, \bibinfo{pages}{17429--17442}
  (\bibinfo{year}{2020}).

\bibitem{rackauckas2020universal}
\bibinfo{author}{Rackauckas, C.} \emph{et~al.}
\newblock \bibinfo{title}{Universal differential equations for scientific
  machine learning}.
\newblock \emph{\bibinfo{journal}{arXiv preprint arXiv:2001.04385}}
  (\bibinfo{year}{2020}).

\bibitem{hoerl1970ridge}
\bibinfo{author}{Hoerl, A.~E.} \& \bibinfo{author}{Kennard, R.~W.}
\newblock \bibinfo{title}{Ridge regression: Biased estimation for nonorthogonal
  problems}.
\newblock \emph{\bibinfo{journal}{Technometrics}}
  \textbf{\bibinfo{volume}{12}}, \bibinfo{pages}{55--67}
  (\bibinfo{year}{1970}).

\bibitem{karniadakis2021physics}
\bibinfo{author}{Karniadakis, G.~E.} \emph{et~al.}
\newblock \bibinfo{title}{Physics-informed machine learning}.
\newblock \emph{\bibinfo{journal}{Nature Reviews Physics}}
  \textbf{\bibinfo{volume}{3}}, \bibinfo{pages}{422--440}
  (\bibinfo{year}{2021}).

\bibitem{lu2021physics}
\bibinfo{author}{Lu, L.} \emph{et~al.}
\newblock \bibinfo{title}{Physics-informed neural networks with hard
  constraints for inverse design}.
\newblock \emph{\bibinfo{journal}{arXiv preprint arXiv:2102.04626}}
  (\bibinfo{year}{2021}).

\bibitem{shin2020convergence}
\bibinfo{author}{Shin, Y.}, \bibinfo{author}{Darbon, J.} \&
  \bibinfo{author}{Karniadakis, G.~E.}
\newblock \bibinfo{title}{On the convergence of physics informed neural
  networks for linear second-order elliptic and parabolic type pdes}.
\newblock \emph{\bibinfo{journal}{arXiv preprint arXiv:2004.01806}}
  (\bibinfo{year}{2020}).

\bibitem{kochkov2021machine}
\bibinfo{author}{Kochkov, D.} \emph{et~al.}
\newblock \bibinfo{title}{Machine learning--accelerated computational fluid
  dynamics}.
\newblock \emph{\bibinfo{journal}{Proceedings of the National Academy of
  Sciences}} \textbf{\bibinfo{volume}{118}}, \bibinfo{pages}{e2101784118}
  (\bibinfo{year}{2021}).

\bibitem{lu2021learning}
\bibinfo{author}{Lu, L.}, \bibinfo{author}{Jin, P.}, \bibinfo{author}{Pang,
  G.}, \bibinfo{author}{Zhang, Z.} \& \bibinfo{author}{Karniadakis, G.~E.}
\newblock \bibinfo{title}{Learning nonlinear operators via deeponet based on
  the universal approximation theorem of operators}.
\newblock \emph{\bibinfo{journal}{Nature Machine Intelligence}}
  \textbf{\bibinfo{volume}{3}}, \bibinfo{pages}{218--229}
  (\bibinfo{year}{2021}).

\bibitem{li2020fourier}
\bibinfo{author}{Li, Z.} \emph{et~al.}
\newblock \bibinfo{title}{Fourier neural operator for parametric partial
  differential equations}.
\newblock \emph{\bibinfo{journal}{arXiv preprint arXiv:2010.08895}}
  (\bibinfo{year}{2020}).

\bibitem{koziel2008space}
\bibinfo{author}{Koziel, S.}, \bibinfo{author}{Cheng, Q.~S.} \&
  \bibinfo{author}{Bandler, J.~W.}
\newblock \bibinfo{title}{Space mapping}.
\newblock \emph{\bibinfo{journal}{IEEE Microwave Magazine}}
  \textbf{\bibinfo{volume}{9}}, \bibinfo{pages}{105--122}
  (\bibinfo{year}{2008}).

\bibitem{bakr2000neural}
\bibinfo{author}{Bakr, M.~H.}, \bibinfo{author}{Bandler, J.~W.},
  \bibinfo{author}{Ismail, M.~A.}, \bibinfo{author}{Rayas-S{\'a}nchez, J.~E.}
  \& \bibinfo{author}{Zhang, Q.-J.}
\newblock \bibinfo{title}{Neural space-mapping optimization for em-based
  design}.
\newblock \emph{\bibinfo{journal}{IEEE Transactions on Microwave Theory and
  Techniques}} \textbf{\bibinfo{volume}{48}}, \bibinfo{pages}{2307--2315}
  (\bibinfo{year}{2000}).

\bibitem{feng2019coarse}
\bibinfo{author}{Feng, F.} \emph{et~al.}
\newblock \bibinfo{title}{Coarse-and fine-mesh space mapping for em
  optimization incorporating mesh deformation}.
\newblock \emph{\bibinfo{journal}{IEEE Microwave and Wireless Components
  Letters}} \textbf{\bibinfo{volume}{29}}, \bibinfo{pages}{510--512}
  (\bibinfo{year}{2019}).

\bibitem{lu2020extraction}
\bibinfo{author}{Lu, L.} \emph{et~al.}
\newblock \bibinfo{title}{Extraction of mechanical properties of materials
  through deep learning from instrumented indentation}.
\newblock \emph{\bibinfo{journal}{Proceedings of the National Academy of
  Sciences}} \textbf{\bibinfo{volume}{117}}, \bibinfo{pages}{7052--7062}
  (\bibinfo{year}{2020}).

\bibitem{koziel2006space}
\bibinfo{author}{Koziel, S.}, \bibinfo{author}{Bandler, J.~W.} \&
  \bibinfo{author}{Madsen, K.}
\newblock \bibinfo{title}{A space-mapping framework for engineering
  optimization—theory and implementation}.
\newblock \emph{\bibinfo{journal}{IEEE Transactions on Microwave Theory and
  Techniques}} \textbf{\bibinfo{volume}{54}}, \bibinfo{pages}{3721--3730}
  (\bibinfo{year}{2006}).

\bibitem{levine2022framework}
\bibinfo{author}{Levine, M.} \& \bibinfo{author}{Stuart, A.}
\newblock \bibinfo{title}{A framework for machine learning of model error in
  dynamical systems}.
\newblock \emph{\bibinfo{journal}{Communications of the American Mathematical
  Society}} \textbf{\bibinfo{volume}{2}}, \bibinfo{pages}{283--344}
  (\bibinfo{year}{2022}).

\bibitem{ren2022physics}
\bibinfo{author}{Ren, P.} \emph{et~al.}
\newblock \bibinfo{title}{Physics-informed deep super-resolution for
  spatiotemporal data}.
\newblock \emph{\bibinfo{journal}{arXiv preprint arXiv:2208.01462}}
  (\bibinfo{year}{2022}).

\bibitem{drygala2022generative}
\bibinfo{author}{Drygala, C.}, \bibinfo{author}{Winhart, B.},
  \bibinfo{author}{di~Mare, F.} \& \bibinfo{author}{Gottschalk, H.}
\newblock \bibinfo{title}{Generative modeling of turbulence}.
\newblock \emph{\bibinfo{journal}{Physics of Fluids}}
  \textbf{\bibinfo{volume}{34}}, \bibinfo{pages}{035114}
  (\bibinfo{year}{2022}).

\bibitem{geneva2020multi}
\bibinfo{author}{Geneva, N.} \& \bibinfo{author}{Zabaras, N.}
\newblock \bibinfo{title}{Multi-fidelity generative deep learning turbulent
  flows}.
\newblock \emph{\bibinfo{journal}{arXiv preprint arXiv:2006.04731}}
  (\bibinfo{year}{2020}).

\bibitem{hesthaven2022reduced}
\bibinfo{author}{Hesthaven, J.~S.}, \bibinfo{author}{Pagliantini, C.} \&
  \bibinfo{author}{Rozza, G.}
\newblock \bibinfo{title}{Reduced basis methods for time-dependent problems}.
\newblock \emph{\bibinfo{journal}{Acta Numerica}}
  \textbf{\bibinfo{volume}{31}}, \bibinfo{pages}{265--345}
  (\bibinfo{year}{2022}).

\bibitem{weinberg1995quantum}
\bibinfo{author}{Weinberg, S.}
\newblock \emph{\bibinfo{title}{The quantum theory of fields}},
  vol.~\bibinfo{volume}{2} (\bibinfo{publisher}{Cambridge university press},
  \bibinfo{year}{1995}).

\bibitem{hou2017iteratively}
\bibinfo{author}{Hou, T.~Y.}, \bibinfo{author}{Hwang, F.-N.},
  \bibinfo{author}{Liu, P.} \& \bibinfo{author}{Yao, C.-C.}
\newblock \bibinfo{title}{An iteratively adaptive multi-scale finite element
  method for elliptic pdes with rough coefficients}.
\newblock \emph{\bibinfo{journal}{Journal of Computational Physics}}
  \textbf{\bibinfo{volume}{336}}, \bibinfo{pages}{375--400}
  (\bibinfo{year}{2017}).

\bibitem{molesky2018inverse}
\bibinfo{author}{Molesky, S.} \emph{et~al.}
\newblock \bibinfo{title}{Inverse design in nanophotonics}.
\newblock \emph{\bibinfo{journal}{Nature Photonics}}
  \textbf{\bibinfo{volume}{12}}, \bibinfo{pages}{659--670}
  (\bibinfo{year}{2018}).

\bibitem{perez2018sideways}
\bibinfo{author}{P{\'e}rez-Arancibia, C.}, \bibinfo{author}{Pestourie, R.} \&
  \bibinfo{author}{Johnson, S.~G.}
\newblock \bibinfo{title}{Sideways adiabaticity: beyond ray optics for slowly
  varying metasurfaces}.
\newblock \emph{\bibinfo{journal}{Optics express}}
  \textbf{\bibinfo{volume}{26}}, \bibinfo{pages}{30202--30230}
  (\bibinfo{year}{2018}).

\bibitem{jin2015finite}
\bibinfo{author}{Jin, J.-M.}
\newblock \emph{\bibinfo{title}{The finite element method in electromagnetics}}
  (\bibinfo{publisher}{John Wiley \& Sons}, \bibinfo{year}{2015}).

\bibitem{svanberg2002class}
\bibinfo{author}{Svanberg, K.}
\newblock \bibinfo{title}{A class of globally convergent optimization methods
  based on conservative convex separable approximations}.
\newblock \emph{\bibinfo{journal}{SIAM journal on optimization}}
  \textbf{\bibinfo{volume}{12}}, \bibinfo{pages}{555--573}
  (\bibinfo{year}{2002}).

\bibitem{liu1989limited}
\bibinfo{author}{Liu, D.~C.} \& \bibinfo{author}{Nocedal, J.}
\newblock \bibinfo{title}{On the limited memory bfgs method for large scale
  optimization}.
\newblock \emph{\bibinfo{journal}{Mathematical programming}}
  \textbf{\bibinfo{volume}{45}}, \bibinfo{pages}{503--528}
  (\bibinfo{year}{1989}).

\bibitem{huber1992robust}
\bibinfo{author}{Huber, P.~J.}
\newblock \bibinfo{title}{Robust estimation of a location parameter}.
\newblock In \emph{\bibinfo{booktitle}{Breakthroughs in statistics}},
  \bibinfo{pages}{492--518} (\bibinfo{publisher}{Springer},
  \bibinfo{year}{1992}).

\bibitem{lakshminarayanan2016simple}
\bibinfo{author}{Lakshminarayanan, B.}, \bibinfo{author}{Pritzel, A.} \&
  \bibinfo{author}{Blundell, C.}
\newblock \bibinfo{title}{Simple and scalable predictive uncertainty estimation
  using deep ensembles}.
\newblock \emph{\bibinfo{journal}{arXiv preprint arXiv:1612.01474}}
  (\bibinfo{year}{2016}).

\bibitem{goodfellow2016deep}
\bibinfo{author}{Goodfellow, I.}, \bibinfo{author}{Bengio, Y.} \&
  \bibinfo{author}{Courville, A.}
\newblock \emph{\bibinfo{title}{Deep learning}} (\bibinfo{publisher}{MIT
  press}, \bibinfo{year}{2016}).

\bibitem{kingma2014adam}
\bibinfo{author}{Kingma, D.~P.} \& \bibinfo{author}{Ba, J.}
\newblock \bibinfo{title}{Adam: A method for stochastic optimization}.
\newblock \emph{\bibinfo{journal}{arXiv preprint arXiv:1412.6980}}
  (\bibinfo{year}{2014}).

\bibitem{innes:2018}
\bibinfo{author}{Innes, M.}
\newblock \bibinfo{title}{Flux: Elegant machine learning with julia}.
\newblock \emph{\bibinfo{journal}{Journal of Open Source Software}}
  (\bibinfo{year}{2018}).

\bibitem{innes2018don}
\bibinfo{author}{Innes, M.}
\newblock \bibinfo{title}{Don't unroll adjoint: Differentiating ssa-form
  programs}.
\newblock \emph{\bibinfo{journal}{arXiv preprint arXiv:1810.07951}}
  (\bibinfo{year}{2018}).

\bibitem{champagne2001fdfd}
\bibinfo{author}{Champagne~II, N.~J.}, \bibinfo{author}{Berryman, J.~G.} \&
  \bibinfo{author}{Buettner, H.~M.}
\newblock \bibinfo{title}{Fdfd: A 3d finite-difference frequency-domain code
  for electromagnetic induction tomography}.
\newblock \emph{\bibinfo{journal}{Journal of Computational Physics}}
  \textbf{\bibinfo{volume}{170}}, \bibinfo{pages}{830--848}
  (\bibinfo{year}{2001}).

\bibitem{sacks1995perfectly}
\bibinfo{author}{Sacks, Z.~S.}, \bibinfo{author}{Kingsland, D.~M.},
  \bibinfo{author}{Lee, R.} \& \bibinfo{author}{Lee, J.-F.}
\newblock \bibinfo{title}{A perfectly matched anisotropic absorber for use as
  an absorbing boundary condition}.
\newblock \emph{\bibinfo{journal}{IEEE transactions on Antennas and
  Propagation}} \textbf{\bibinfo{volume}{43}}, \bibinfo{pages}{1460--1463}
  (\bibinfo{year}{1995}).

\bibitem{rpestourie_payel79_2023}
\bibinfo{author}{Pestourie, R.} \& \bibinfo{author}{Das, P.}
\newblock \bibinfo{title}{payel79/peds: publish
  https://doi.org/10.5281/zenodo.8342595} (\bibinfo{year}{2023}).

\end{thebibliography}


\begin{thebibliography}{10}

\bibitem{pestourie2020active}
Rapha{\"e}l Pestourie, Youssef Mroueh, Thanh~V Nguyen, Payel Das, and Steven~G Johnson.
\newblock Active learning of deep surrogates for pdes: application to metasurface design.
\newblock {\em npj Computational Materials}, 6(1):1--7, 2020.

\bibitem{koziel2008space}
Slawomir Koziel, Qingsha~S Cheng, and John~W Bandler.
\newblock Space mapping.
\newblock {\em IEEE Microwave Magazine}, 9(6):105--122, 2008.

\bibitem{bakr2000neural}
Mohamed~H Bakr, John~W Bandler, Mostafa~A Ismail, Jos{\'e}~E Rayas-S{\'a}nchez, and Qi-Jun Zhang.
\newblock Neural space-mapping optimization for em-based design.
\newblock {\em IEEE Transactions on Microwave Theory and Techniques}, 48(12):2307--2315, 2000.

\bibitem{feng2019coarse}
Feng Feng, Jianan Zhang, Wei Zhang, Zhihao Zhao, Jing Jin, and Qi-Jun Zhang.
\newblock Coarse-and fine-mesh space mapping for em optimization incorporating mesh deformation.
\newblock {\em IEEE Microwave and Wireless Components Letters}, 29(8):510--512, 2019.

\bibitem{zhu2016novel}
Lin Zhu, Qijun Zhang, Kaihua Liu, Yongtao Ma, Bo~Peng, and Shuxia Yan.
\newblock A novel dynamic neuro-space mapping approach for nonlinear microwave device modeling.
\newblock {\em IEEE Microwave and Wireless Components Letters}, 26(2):131--133, 2016.

\bibitem{srivastava2014dropout}
Nitish Srivastava, Geoffrey Hinton, Alex Krizhevsky, Ilya Sutskever, and Ruslan Salakhutdinov.
\newblock Dropout: a simple way to prevent neural networks from overfitting.
\newblock {\em The journal of machine learning research}, 15(1):1929--1958, 2014.

\bibitem{holloway2011characterizing}
Christopher~L Holloway, Edward~F Kuester, and Andrew Dienstfrey.
\newblock Characterizing metasurfaces/metafilms: the connection between surface susceptibilities and effective material properties.
\newblock {\em IEEE Antennas Wirel. Propag. Lett.}, 10:1507--1511, 2011.

\bibitem{branicki2013fundamental}
Michal Branicki and Andrew~J Majda.
\newblock Fundamental limitations of polynomial chaos for uncertainty quantification in systems with intermittent instabilities.
\newblock {\em Communications in mathematical sciences}, 11(1):55--103, 2013.

\bibitem{jentzen2018proof}
Arnulf Jentzen, Diyora Salimova, and Timo Welti.
\newblock A proof that deep artificial neural networks overcome the curse of dimensionality in the numerical approximation of kolmogorov partial differential equations with constant diffusion and nonlinear drift coefficients.
\newblock {\em arXiv preprint arXiv:1809.07321}, 2018.

\bibitem{hornik1989multilayer}
Kurt Hornik, Maxwell Stinchcombe, and Halbert White.
\newblock Multilayer feedforward networks are universal approximators.
\newblock {\em Neural networks}, 2(5):359--366, 1989.

\bibitem{ismailov2023three}
Vugar~E Ismailov.
\newblock A three layer neural network can represent any multivariate function.
\newblock {\em Journal of Mathematical Analysis and Applications}, 523(1):127096, 2023.

\bibitem{kidger2020universal}
Patrick Kidger and Terry Lyons.
\newblock Universal approximation with deep narrow networks.
\newblock In {\em Conference on learning theory}, pages 2306--2327. PMLR, 2020.

\bibitem{woodworth2020kernel}
Blake Woodworth, Suriya Gunasekar, Jason~D Lee, Edward Moroshko, Pedro Savarese, Itay Golan, Daniel Soudry, and Nathan Srebro.
\newblock Kernel and rich regimes in overparametrized models.
\newblock In {\em Conference on Learning Theory}, pages 3635--3673. PMLR, 2020.

\end{thebibliography}

\end{document}


\maketitle

\noindent \normalsize{$^{1}$ School of Computational Science and Engineering, Georgia Institute of Technology, Atlanta, GA 30332, USA} \\
\normalsize{$^{2}$ IBM Research AI, IBM Thomas J Watson Research Center,  Yorktown Heights, NY 10598, USA}\\
\normalsize{$^{3}$ MIT-IBM Watson AI Lab, Cambridge, MA 02139, USA}\\
\normalsize{$^{4}$ MIT, Cambridge, MA 02139, USA}\\
\normalsize{$^\ast$Correspondence to: rpestourie3@gatech.edu; daspa@us.ibm.com.}

\tableofcontents

\section{Optimal mixing weights for Fourier($16$), Fourier($25$), Fisher($16$), and Fisher($25$)}

We show the optimal mixing coefficients for Fourier($16$), Fourier($25$), Fisher($16$), and Fisher($25$) in Table~1. Each column corresponds to a model in the ensemble. The weights for the diffusion equation are smaller because the low-fidelity is more accurate. The weights for the nonlinear reaction--diffusion equation are $\approx5\times$ greater because the neural generator has the stronger impact of correcting for the nonlinear effects.
\begin{table}[ht]\label{tab:cw}
\begin{tabular}{l|r|r|r|r|r|}
\cline{2-6}
                                  & \multicolumn{1}{l|}{Model 1} & \multicolumn{1}{l|}{Model 2} & \multicolumn{1}{l|}{Model 3} & \multicolumn{1}{l|}{Model 4} & \multicolumn{1}{l|}{Model 5} \\ \hline
\multicolumn{1}{|l|}{Fourier(16)} & 0.0931965                    & 0.0859197                    & 0.0904296                    & 0.0877423                    & 0.0900292                    \\ \hline
\multicolumn{1}{|l|}{Fourier(25)} & 0.0990645                    & 0.10286                      & 0.102562                     & 0.101054                     & 0.103778                     \\ \hline
\multicolumn{1}{|l|}{Fisher(16)}  & 0.451373                     & 0.448773                     & 0.452949                     & 0.455903                     & 0.452241                     \\ \hline
\multicolumn{1}{|l|}{Fisher(25)}  & 0.471411                     & 0.472364                     & 0.467987                     & 0.468141                     & 0.467343                     \\ \hline
\end{tabular}
\caption{Optimal mixing coefficients for Fourier($16$), Fourier($25$), Fisher($16$), and Fisher($25$). Each column corresponds to a model in the ensemble. The weights for the diffusion equation are smaller because the low-fidelity is more accurate. The weights for the nonlinear reaction--diffusion equation are $\approx5\times$ greater because the neural generator has the stronger impact of correcting for the nonlinear effects.}
\end{table}

\section{Implementation details of PEDS and baselines for Maxwell($10$)}
The generator neural network of PEDS has two hidden layers with 256 nodes and relu activation functions, and outputs a flattened version of the coarse geometry of dimension 1100 with a hardtanh activation function ($hardtan(x)=max(min(x, 1), 0)$). The network that outputs the variance of the models, takes the generated coarse geometry as input, has 3 hidden layers with relu activation functions and outputs a scalar with a relu activation function. 
The corresponding baseline, which is a neural-network only (NN-only) method, was chosen to be as close as possible to PEDS architecture, it replaces the coarse solver with a fully connected layer, and outputs two scalars with a tanh activation function. Note that it does not have the information of the downsampled structure.
The mapping neural network of the input SM implementation has two hidden layers with 256 nodes and relu activation functions, and outputs  the coarse geometry parameters of dimension 10 with a hardtanh activation function. The variance network is similar to PEDS except that the inputs are the SM output geometry parameters.
The batch size was set to 64 and the learning rate to $10^{-3}$. Every training went through 10 epochs.

\section{Active learning implementation details}
 The active learning training~\cite{pestourie2020active} used the following parameters $n_\mathrm{init}= 256$, $T=8$, $M=4$, and K took powers of 2 ranging from $2^6$ to $2^{16}$. 

\section{Parallelization}
In order to accelerate the training of the surrogate model, we parallelized the training at the batch loop level. For ensemble learning, we used 320 computing units which were split into 5 groups (one group per model in the ensemble) of 64 computing units. With a batch size of 64, each worker evaluates the surrogate only once per batch loop (a batch size of a multiple of 64 would work well too).

\section{SM baseline}
Our PEDS has similarities with input space mapping (SM)~\cite{koziel2008space}, especially neural SM~\cite{bakr2000neural} and coarse mesh/fine SM~\cite{feng2019coarse}, where the input of a fine solver is mapped into the input of a coarse solver. However, SM uses the same parameterization $p$ (e.g. the widths of the holes) for the fine solver and the coarse solver, so the input dimension and the output dimension of the neural generator are equal. In contrast, 
PEDS uses a much richer coarse-geometry input (a grid of material values, whose dimensionality is different in the coarse and fine geometries) and can therefore incorporate more geometric and physical inductive biases, such as symmetries and the downsampled structure. For comparison, we trained an input SM baseline model, which is a combination of neural and coarse mesh/fine mesh SM~\cite{zhu2016novel, feng2019coarse}. In this model, the NN is learning a mapping that creates modified geometry parameters $p$ which are combined with an intuition-based geometry and then fed to the coarse solver. Since the coarsified parameterized geometry is implemented via sub-pixel averaging, this function is differentiable, so the gradient can backpropagate all the way to the mapping NN.

The baseline corresponding to the previous ``space-mapping''~(SM) \cite{bakr2000neural, zhu2016novel, feng2019coarse} approach combines a coarse Maxwell solver with a NN  transforming only a low-dimensional parameterization of the fine geometry to a similar low-dimensional parameterization of the coarse geometry.

\begin{figure}[ht]
    \centering
    \includegraphics[width=\textwidth]{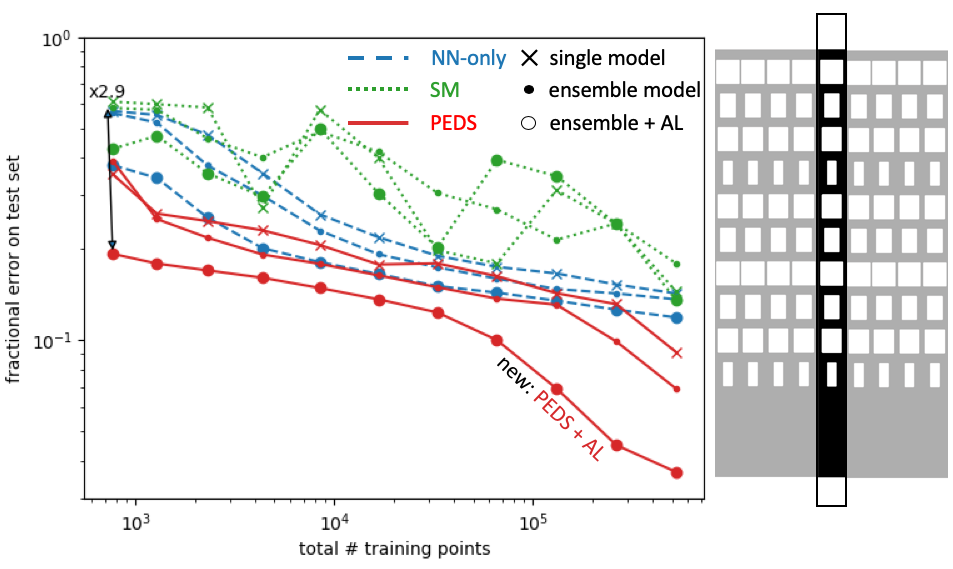}
    \caption{(Left) Fractional error (FE) on the test set with a coarse resolution of 10: PEDS outperforms the other models significantly when combined with active learning (AL). SM performs poorly compared to PEDS and does not gain much accuracy from ensembling nor from active learning. (Right) Geometry of the unit cell of the surrogate model. Each of the 10 air holes have independent widths, the simulation is performed with periodic boundary conditions on the long sides, the incident light comes from the bottom and the complex transmission is measured at the top of the geometry.}
    \label{fig:resultfigure10}
\end{figure}

\begin{figure}[ht]
    \centering
    \includegraphics[width=\textwidth]{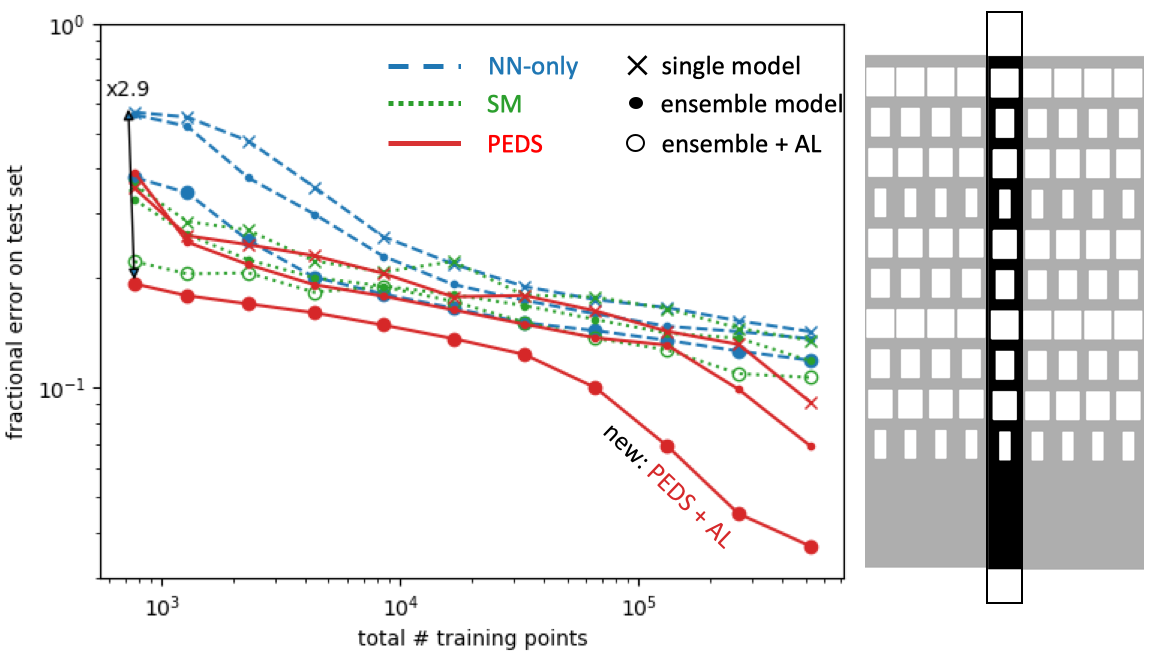}
    \caption{(Left) Fractional error (FE) on the test set with a coarse resolution of 20: PEDS outperforms the other models significantly when combined with active learning (AL). SM performs poorly compared to PEDS and does not gain much accuracy from ensembling nor from active learning. SM performs similarly to the neural network baseline at this resolution. (Right) Geometry of the unit cell of the surrogate model. Each of the 10 air holes have independent widths, the simulation is performed with periodic boundary conditions on the long sides, the incident light comes from the bottom and the complex transmission is measured at the top of the geometry.}
    \label{fig:resultfigure20}
\end{figure}

As shown in Fig.~\ref{fig:resultfigure10}, SM performs significantly worse than PEDS, and does not gain much accuracy from ensembling nor from active learning, and is even worse than the baseline NN. We found that SM can perform comparably to the baseline NN if the coarse-solver resolution is doubled to 20, shown in Fig.~\ref{fig:resultfigure20}, at the expense of $\approx 10\times$ more computational effort.

\section{Coarse solver and gradient}

In the present work, the coarse solver is similar to the fine solver except that it uses a much coarser resolution of 10, which corresponds to a resolution of less than 5 pixels per wavelength in the worst case, instead of 40 for the fine model. The symmetry action was a simple mirror symmetry, implemented by averaging of the geometry with its mirror flip.

\section{Implementation details of PEDS and baselines for Fourier/Fisher($16$) and Fourier/Fisher($25$)}
The generator neural network of PEDS has two hidden layers with 128 nodes and relu activation functions, and outputs a flattened version of the coarse geometry of dimension 16 and 25 with a hardtanh activation function ($hardtan(x)=max(min(x, 1), 0)$), for 16 and 25 input parameters, respectively. Each hidden layer is followed by a dropout~\cite{srivastava2014dropout} of rate 50\%.
The corresponding baseline, which is a neural-network only (NN-only) method, was chosen to be as close as possible to PEDS architecture, it replaces the coarse solver with a fully connected layer, and outputs two scalars with a tanh activation function. Note that it does not have the information of the downsampled structure.
The batch size was set to 64 and the learning rate to $5 10^{-5}$. Every training went through 200 epochs. And the dataset contains 1088 points ($\approx 10^3$).

\section{Ablation study} 
Next, we show results of ablation experiments in order to understand the effect of mixing the generated structure with a downsampled structure. Specifically, we performed an ablation study on an AL ensemble model in the low-data regime (1280 training points); results are shown in Table~\ref{tab:ablation}. The edge cases of using only the downsampled structure with the low-fidelity solver (Table~\ref{tab:ablation}, coarsified only) performs the worst (124\% error with respect to the high-fidelity solver), corresponding to $w=0.0$ in \eqref{model}. Conversely, using the NN generator only (Table~\ref{tab:ablation}, generator only), corresponding to $w=1.0$ in \eqref{model}, is still about 15\% worse (0.20 error) than using adaptive mixing $0 < w < 1$ (Table~\ref{tab:ablation}, PEDS). Imposing mirror symmetry, via $P[G] = (G + \mbox{mirror image})/2$ in \eqref{model} (Table~\ref{tab:ablation}, PEDS with symmetry), did not improve the accuracy of the model in this case (but is a useful option in general, since symmetry may have a larger effect on the physics in other applications).

\begin{table}[h!]
\centering
\begin{tabular}{lll}
\hline
Generative model for low-fidelity geometry & FE on test set & PEDS improvement \\\hline
$w = 0.0$ (coarsified only)                      & 1.24      & 86\%             \\
$w = 1.0$ (generator only)                   & 0.20      &           15\%  \\
PEDS with symmetry
&  0.18            &     5\%    \\
PEDS
&  0.17            &     ---    \\
\hline
\end{tabular}
\caption{Ablation study of PEDS with ensembling and active learning for 1280 training points, showing the impact of mixing generated and coarsified geometries, as well as imposing symmetry.}
\label{tab:ablation}
\end{table}

\begin{figure}
    \centering
    \includegraphics[width=\textwidth]{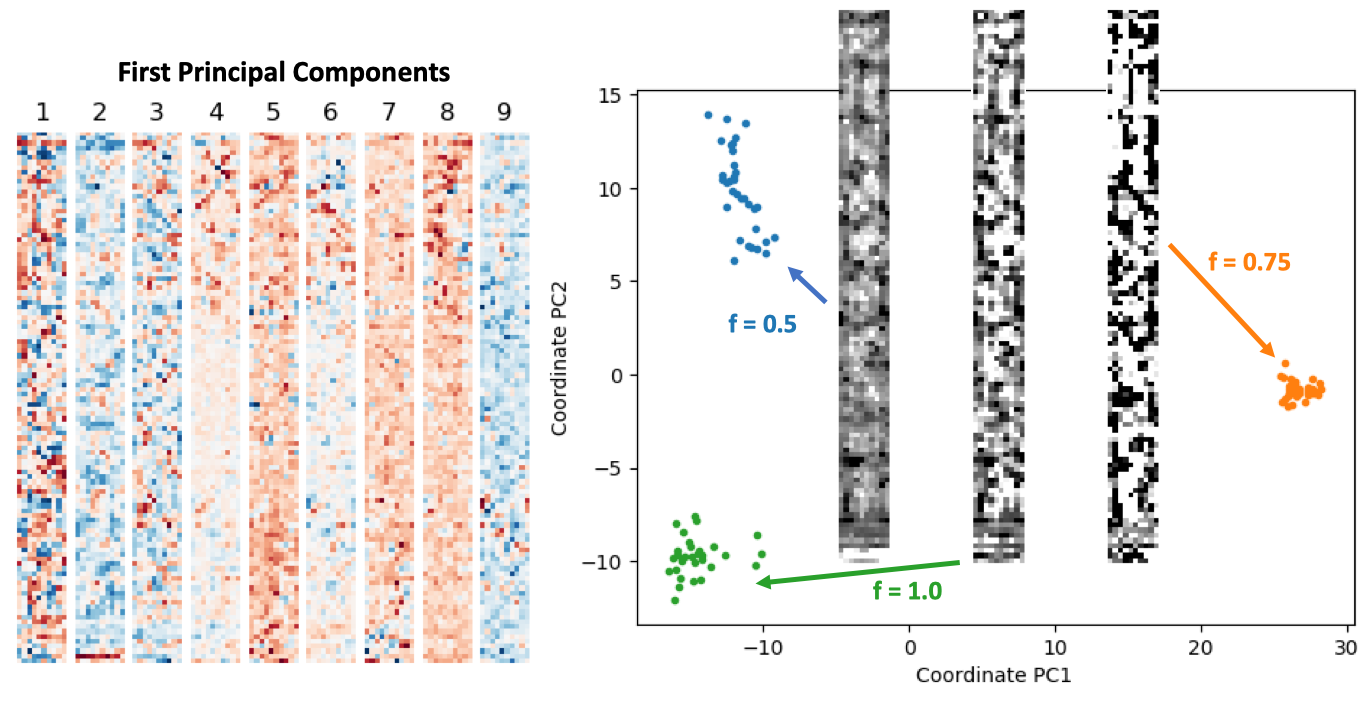}
    \caption{ (Left) First 9 principal components which explain most of the variation in the complex transmission. (Right) Coordinate of randomly generated structures on the two first principal components. Clusters can clearly discriminate the input geometries ($f=0.5$ in blue, $f=0.75$ in orange, $f=1.0$ in green). (Insets) Example generated geometries corresponding to the three frequencies of the surrogate model. The generated geometry is smoothest for the smallest frequency.}
    \label{fig:generatedstudy}
\end{figure}

\section{Analysis of generated geometries} Because the trained PEDS model includes a NN that generates ``equivalent'' coarse-grained geometries to the input structure, it is interesting to analyze these geometries and potentially extract physical insights.

\section{Frequency dependence }
The neural network generates structures that are qualitatively different as a function of the input frequency (Fig.~\ref{fig:generatedstudy}, right insets). As might be expected on physical grounds (e.g. effective-medium theory~\cite{holloway2011characterizing}), the lowest frequency (longer wavelengths) corresponds to the smoothest generated structures, because the wavelength sets the minimum relevant lengthscale for wave scattering. To help quantify this, we performed a principal components analysis (PCA) of $\mathrm{generator}_\mathrm{NN}(p)$ for $10^5$ uniform random $p$ values (including random frequency). We show the first few principal components in Fig.~\ref{fig:generatedstudy}~(left). The first and second components explain 67\% and 13\% of the variation, respectively. We show in Fig.~\ref{fig:generatedstudy}~(right) that the coordinates of the first two components are sufficient to classify generated geometries according to the input frequency.

\section{Scattering richness } To explore the effect of additional scattering physics produced by multiple layers of holes, we generated coarse geometries for different numbers of layers (equivalently, fixing the parameters of the ``deleted'' layers to zero).  We then decomposed the resulting $\mathrm{generator}_\mathrm{NN}(p)$ into the PCA components from above. As we increase the number of layers, the average coordinates of some principal components monotonically increase in magnitude. Since we know that more layers contain more scattering richness, the corresponding principal components geometries provides some geometrical insight into how scattering richness translates into the generated structure. From our analysis of generated structures for the smallest frequency, the first principal component geometry clearly contributes to scattering richness, with an average coordinate (across $10^3$ generated structures) increasing -11 to 26 as the number of layers goes from 1 to 9.

\section{\edit{Comparison to mainstream surrogate models}}
\edit{In the following section we show that the method should be matching the complexity of the problem. Neural network surrogates are not one model-fit-all type of surrogates, but they are suitable for more complex models. For example, the Maxwell surrogate is the most complex with 10 independent geometry parameters~\cite{pestourie2020active}, whereas the diffusion equation is smoothing and therefore low-rank. For the more complex case of Maxwell’s equations, PEDS outperforms NN-only, polynomial chaos expansion (Poly Chaos), radial basis function interpolation (RadialBasis) and gaussian processes (GP), by 66\%, 74\%, 44\%, and 49\%, respectively (Table~\ref{tab:tradsur}). For the simpler cases of the diffusion equation and the reaction-diffusion equation with a small nonlinear term, more traditional surrogates such as Poly Chaos, RadialBasis and GP are competitive against neural networks. Polynomial chaos expansion is performing best on the simple cases and worst worst performer for the more complex surrogate, this is consistent with their underperformance as nonlinearity increases~\cite{branicki2013fundamental}. When many points are needed to be accurate as in the Maxwell case where PEDS' outperforming accuracy of $19\%$ is still much higher than the fabrication error, neural networks training scales linearly with the number of training points; in contrast, radial basis function, gaussian processes, or polynomial chaos expansion have $O(N^3)$
complexity in the worst case scenario where N is the number of training points. However, they can be significantly improved to $O(N M^2)$---where M is a hyperparameter of the algorithm–by exploiting the mathematical structure of problem. On one hand, the neural networks in PEDS reduce the curse of dimensionality~\cite{jentzen2018proof}, they perform nonlinear fits  which makes them suitable for complex applications. On the other hand, the choice of low-fidelity solver controls how much field knowledge is included inside PEDS, and adds a strong regularization towards physical solutions that makes it more data efficient. These features makes them suitable for complex applications.}

\begin{table}[h]
\includegraphics[width=\linewidth]{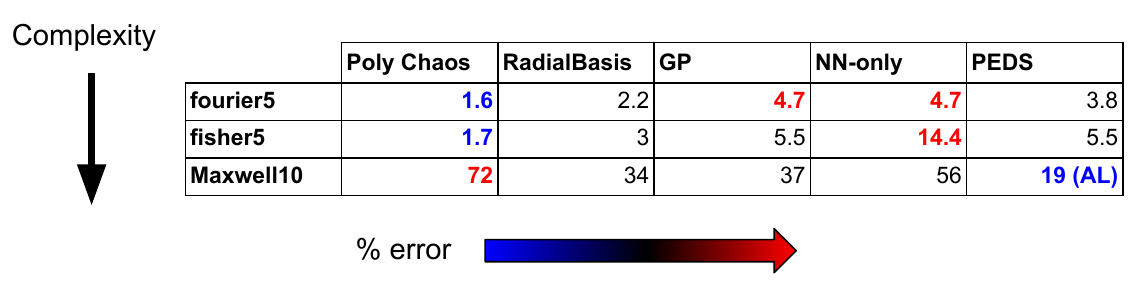}
\caption{\edit{Comparison of PEDS and NN-only models against traditional surrogate models: as the complexity of the generative process increases, neural network surrogates perform best, but traditional surrogates are competitive in simpler cases.}}
\label{tab:tradsur}
\end{table}

\section{Fractional error}
For evaluation, we use the fractional error $FE$ between two vectors of complex values $\bold{u}_\mathrm{estimate}$ and $\bold{v}_\mathrm{true}$, which is \begin{equation}
    FE = \frac{|\bold{u}_\mathrm{estimate}-\bold{v}_\mathrm{true}|}{|\bold{v}_\mathrm{true}|}
\end{equation} where $|\cdot|$ is the L$2$-norm for complex vectors.

\section{Code}
The code was implemented in Julia language version 1.6, using MPI.jl for parallelization with MPI, Flux.jl for the neural network training framework, ChainRules.jl for custom differentiation rules, and Zygote.jl for other automatic differentiation.

\section{Runtime benchmarking}
Runtime benchmarks were computed on 3.5 GHz 6-Core Intel Xeon E5.

\section{General robustness}
We further studied the general robustness of PEDS---the generalization error---in the most difficult case of Maxwell’s equations. We consider models without ensembling and without active learning to single out the effect of PEDS in comparison to Neural Networks (NN-only) and a baseline of predicting the mean. We study the robustness on random data splits and stratified splits of the test set (Table~2). We report that PEDS' error is 5x more robust to random splits in the test set, and PEDS improvement compared to the baseline is robust to test set splits. On random splits, the NN-only and mean prediction baseline show errors that vary by about 1\% relative to the error on the full test set. PEDS' error varies 5 times less and  is more robust to the choice of test set. On stratified test sets, with data points with high absolute transmission in one set and lower absolute transmission in the other, we see that it is harder to predict lower absolute transmission accurately across the models, but PEDS is always performing much better than the two baselines by a factor of about two at least.

\begin{table}[]\label{tab:robust}
\begin{tabular}{|l|l|l|l|}
\hline
\textbf{}                                            & PEDS        & NN-only     & Mean prediction \\ \hline
Fractional Error (FE) (in \%) on test set            & 28.33       & 53.65       & 95.58           \\ \hline
Improvement with respect to PEDS (times)             & N/A         & 1.89        & 3.40            \\ \hline
FE (in \%) with random splits (set 1/set 2)          & 28.28/28.38 & 54.67/52.49 & 97.38/95.68     \\ \hline
Improvement with respect to PEDS (times)             & N/A         & 1.93/1.85   & 3.44/3.37       \\ \hline
FE  (in \%) with stratified split (set high/set low) & 20.19/36.35 & 46.02/62.28 & 92.35/101.87    \\ \hline
Improvement with respect to PEDS (times)             & N/A         & 2.28/1.71   & 4.57/2.80       \\ \hline
\end{tabular}
\caption{General robustness on random data splits and stratified splits of the test set for models without ensembling and without active learning on the Maxwell(10) surrogate.}
\end{table}

\edit{\section{Exemplar comparison of PEDS against a middle-fidelity solver}
We conducted the numerical comparison between the middle-fidelity solver and PEDS in the case of the surrogate model for Maxwell’s equations. To showcase the benefit of PEDS compared to a middle-fidelity solver, we gradually changed the resolution of the middle fidelity solver from 38 to 30 and computed the fractional error on the test set. In Fig. R1, we can see that the middle-fidelity solver with a resolution of 36 has an error 20\%, which corresponds approximately to PEDS performance when training with only $\approx$1000 data points (19\%). The middle fidelity solver with a resolution of 36 takes 180 ms to evaluate. In contrast, PEDS with a coarse resolution of 10 and including the neural network inference takes 5 ms which is still 36 times faster than the middle-fidelity solver. We conclude that using PEDS has a clear advantage over using a middle-fidelity solver with respect to evaluation time. }

\begin{figure}
    \centering
    \includegraphics[width=\textwidth]{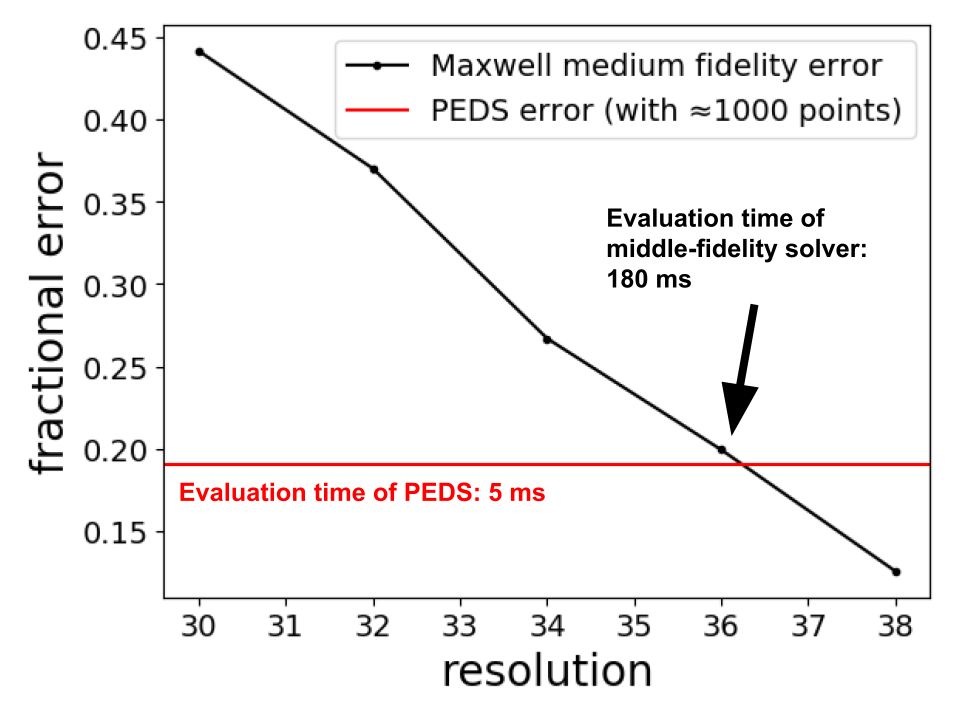}
    \caption{\edit{Comparison of error between various middle-fidelity solver and PEDS}}
    \label{fig:enter-label}
\end{figure}

\edit{\section{Universal Approximation Theorem for PEDS}
The universal approximation theorem only offers a proof of existence of an arbitrarily accurate approximation using PEDS. The existence does not give much practical advice regarding the choices of low-fidelity solver, neural network architecture, dataset etc which impact performance most. However, it does give a strong intuition why PEDS’ accuracy is not limited by that of the low fidelity solver.
When the implicit function theorem applies to the low-fidelity solver function (when the Hessian is invertible), there exists an inverse function of the low-fidelity solver locally. Then the universal approximation theorem of feedforward neural networks with rectified linear unit activation functions~\cite{hornik1989multilayer} can be applied to this inverse function. Recent work on the universal approximation theorem~\cite{ismailov2023three, kidger2020universal} has extended the class of function that can be approximated to discontinuous functions.
In PEDS, the neural network attempts to learn the inverse function of the low-fidelity solver which is defined when the conditions of  implicit function theorem are valid; these conditions may not be met everywhere unless there is some regularization. Under some regularization, the hessian becomes invertible and we get  the pseudo-inverse and not the inverse. In the case of PEDS, the neural network approximates this pseudo inverse with a built-in regularization such as in the kernel regime of stochastic gradient descent~\cite{woodworth2020kernel}.}

\bibliographystyle{unsrt} 
\bibliography{refs.bib}